\documentclass[authoryear,preprint,12pt]{elsarticle}

\usepackage{amssymb}

\usepackage{graphicx}
\usepackage{amssymb}
\usepackage{subcaption}
\graphicspath{{Pics/}}
\usepackage{soul}
\usepackage{xcolor}
\usepackage{bm}
\usepackage{amsmath}
\usepackage{tabularx}
\usepackage{amsthm}
\usepackage[]{hyperref}
\usepackage{apalike}

\newtheorem{definition}{Definition}[section]

\DeclareMathOperator*{\E}{\mathbb{E}}

\journal{Neural Networks}

\begin{document}

\begin{frontmatter}



\title{Investigating the interaction between gradient-only line searches and different activation functions}


\author[label1]{Dominic Kafka}
\author[label2]{Daniel N. Wilke}

\address[label1]{dominic.kafka@gmail.com}
\address[label2]{wilkedn@gmail.com}

\address{Centre for Asset and Integrity Management (C-AIM), \\
	Department of Mechanical and Aeronautical Engineering, \\
	University of Pretoria, Pretoria, South Africa.}

\begin{abstract}
Gradient-only line searches (GOLS) adaptively determine step sizes along search directions for discontinuous loss functions resulting from dynamic mini-batch sub-sampling in neural network training. Step sizes in GOLS are determined by localizing Stochastic Non-Negative Associated Gradient Projection Points (SNN-GPPs) along descent directions. These are identified by a sign change in the directional derivative from negative to positive along a descent direction. Activation functions are a significant component of neural network architectures as they introduce non-linearities essential for complex function approximations. The smoothness and continuity characteristics of the activation functions directly affect the gradient characteristics of the loss function to be optimized. Therefore, it is of interest to investigate the relationship between activation functions and different neural network architectures in the context of GOLS. We find that GOLS are robust for a range of activation functions, but sensitive to the Rectified Linear Unit (ReLU) activation function in standard feedforward architectures. The zero-derivative in ReLU's negative input domain can lead to the gradient-vector becoming sparse, which severely affects training. We show that implementing architectural features such as batch normalization and skip connections can alleviate these difficulties and benefit training with GOLS for all activation functions considered. 
\end{abstract}




\end{frontmatter}


\section{Introduction}
\label{sec_intro}


The recent introduction of Gradient-Only Line Searches (GOLS) \citep{Kafka2019jogo} has enabled learning rates to be determined automatically in the discontinuous loss functions of neural networks training with dynamic mini-batch sub-sampling (MBSS). The discontinuous nature of the dynamic MBSS loss is a direct result of successively sampling different mini-batches from the training data at every function evaluation, introducing a {\it sampling error} \citep{Kafka2019jogo}. To determine step sizes, GOLS locates {\it Stochastic Non-Negative Associated Gradient Projection Points} (SNN-GPPs), manifesting as sign changes from negative to positive in the directional derivative along a descent direction. This allows GOLS to strike a balance between the benefits of training using dynamic MBSS, such as 1) increasing the training algorithm's exposure to training data \citep{Bottou2010} as well as 2) overcoming local minima \citep{Saxe2013,Dauphin2014,Goodfellow2015,Choromanska2015}; and the ability to localize optima in discontinuous loss functions \citep{Kafka2019b}. This is in contrast to minimization line searches \citep{Arora2011}, which find false local minima induced by sampling error discontinuities. These have found to be uniformly spread along the descent direction, rendering minimization line searches ineffective for determining representative step sizes \citep{Kafka2019b,Kafka2019jogo}. 

Previous work has shown, that the {\it Gradient-Only Line Search that is Inexact} (GOLS-I) is capable of determining step sizes for training algorithms beyond stochastic gradient descent (SGD) \citep{Robbins1951}, such as Adagrad \citep{Duchi2011}, which incorporates approximate second order information \citep{Kafka2019jogo}. GOLS-I has also been demonstrated to outperform probabilistic line searches \citep{Mahsereci2017a}, provided mini-batch sizes are not too small ($<50$ for investigated problems) \citep{Kafka2019}. The gradient-only optimization paradigm has recently also shown promise in the construction of approximation models to conduct line searches \citep{Chae2019}.


Some of the most important factors governing the nature of the computed gradients are: 1) The neural network architecture, 2) the activation functions (AFs) used within the architecture, 3) the loss function implemented, and 4) the mini-batch size used to evaluate the loss, to name a few. In this study, we concentrate on the influence of activation functions (AFs) on training performance of GOLS for different neural network architectures. The stability characteristics of neural network losses with different activation functions have been extensively studied \citep{Liu2017}. Activation functions have a direct influence on the loss surface properties of a neural network, and by extension, dictate the nature of the gradients used in GOLS. Therefore, we empirically study how different activation functions affect training when implementing GOLS to determine learning rates. We also consider the effect of architectural features such as batch normalization \citep{Ioffe2015} and skip connections \citep{He2016} on training architectures with different AFs using GOLS.

The AFs considered in this study can be split primarily into two categories, namely:
\begin{enumerate}
	\item The saturation class \citep{Xu2016}: Including Sigmoid \citep{Han1995}, Tanh \citep{Bergstra2009} and Softsign \citep{Karlik2015}; and
	\item The sparsity class: Including ReLU \citep{Glorot2011}, leaky ReLU \citep{Maas2013} and ELU \citep{Clevert2016}.
\end{enumerate}

{
	
	\begin{figure}[h]
		\centering
		\begin{subfigure}{.4\textwidth}
			\centering 
			\includegraphics[width=0.9\linewidth]{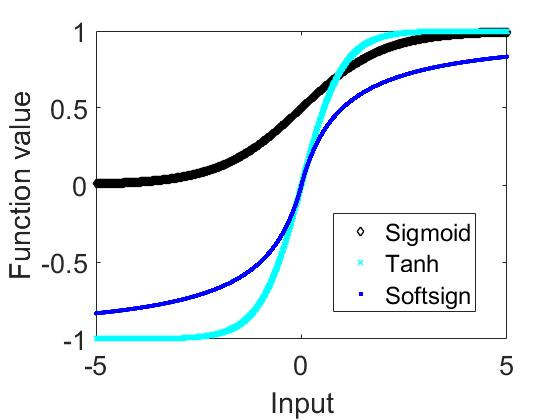}
			\caption{Saturation class function values}
			\label{fig_acts_g1_f}
		\end{subfigure}%
		\begin{subfigure}{.4\textwidth}
			\centering
			\includegraphics[width=0.9\linewidth]{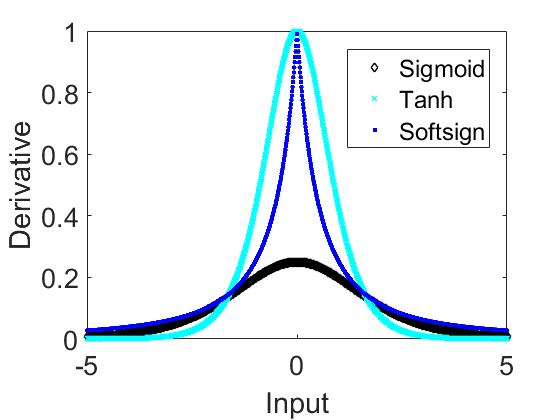}
			\caption{Saturation class derivatives}
			\label{fig_acts_g1_g}
		\end{subfigure}%
		
		\begin{subfigure}{.4\textwidth}
			\centering 
			\includegraphics[width=0.9\linewidth]{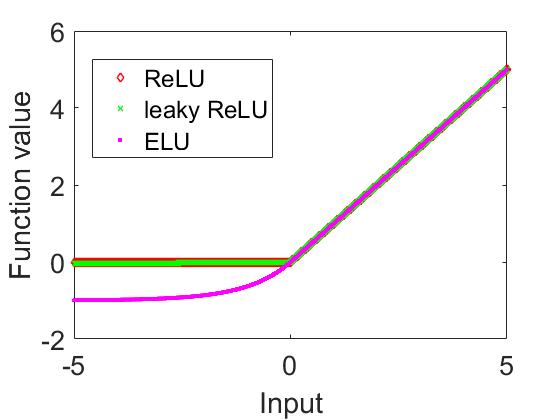}
			\caption{Sparsity class function values}
			\label{fig_acts_g2_f}
		\end{subfigure}%
		\begin{subfigure}{.4\textwidth}
			\centering
			\includegraphics[width=0.9\linewidth]{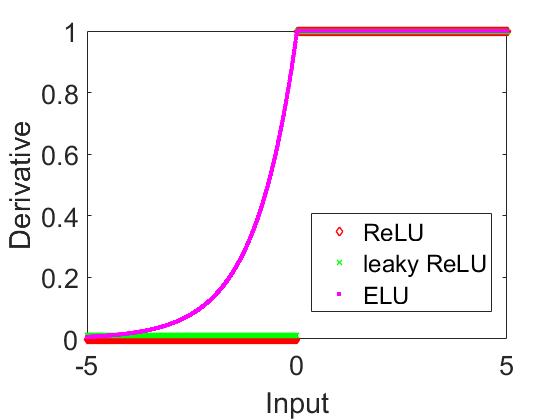}
			\caption{Sparsity class derivatives}
			\label{fig_acts_g2_g}
		\end{subfigure}%

		\caption{(a,c) Function value and (b,d) derivatives of activation functions considered in our investigations. These are grouped together into (a,b) saturation and (c,d) sparsity classes respectively. The primary difference between saturation and sparsity classes are the derivatives in the positive input domain. The saturation class is characterized by derivatives that tend towards zero as input tends toward $+\infty$. Conversely, the sparsity class, is characterized by unit derivatives that spread over all of the positive input domain. This gives the sparsity class AFs behaviour characteristics that approximate a "switch", being either "on" or "off".}
		\label{fig_acts}
	\end{figure}
	
	The respective function values and derivatives of both classes are shown in Figure~\ref{fig_acts} over an input domain of $[-5,5]$. The saturation class is predominantly characterized by derivatives that tend to zero, as the inputs tend to $\pm \infty$. The function values begin either at 0 (as for Sigmoid) or -1 (Tanh and Softsign) have an upper limit of 1. Often, function values and derivatives of the saturation class are smooth and continuous. The outlier in the saturation AFs chosen for this study is Softsign, which has a derivative that is continuous, but not smooth where the input is 0. The derivative characteristics of the Sigmoid AF are also notable among the saturation class AFs. The maximum derivative value of Sigmoid is $0.25$, which is a factor of 4 lower than those of Tanh and Softsign with unit derivatives at the origin.
	
	The original sparsity AF, ReLU, was introduced to recreate the sparseness and switching behaviour observed in neuroscientific studies, in artificial neural networks \citep{Glorot2011}. ReLU is characterized by having an output of zero in the negative input domain, and a linear output with unit gradient in the positive domain. This makes the function values of ReLU continuous and non-smooth, while the derivative is step-wise discontinuous at input 0. As with ReLU, the sparsity class is characterized by having linear outputs in the positive input domain, while the derivatives approximate zero as the negative input domains tend to $-\infty$. However, the derivatives in the positive input domains are always 1, which is a critical difference to the saturation class of AFs. The leaky ReLU AF relaxes the absolute sparsity of ReLU by allowing a small constant derivative in the negative input domain. However, the non-smooth function value and the discontinuous derivative properties of ReLU are maintained. The ELU AF is a further modification that enforces smoothness in the function value and continuity in the derivative. However, the derivative remains non-smooth. The formulations of leaky ReLU and ELU are both claimed to improve training performance over ReLU \citep{Clevert2016}.
	
}

\begin{figure}[h!]
	\centering
	\begin{subfigure}{.33\textwidth}
		\centering 
		\includegraphics[width=0.99\linewidth]{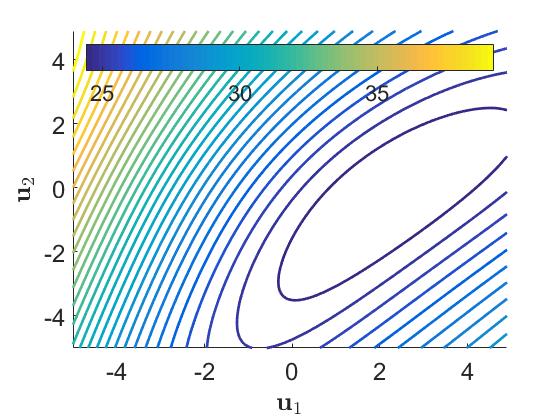}
		\caption{Sigmoid}
	\end{subfigure}%
	\begin{subfigure}{.33\textwidth}
		\centering 
		\includegraphics[width=0.99\linewidth]{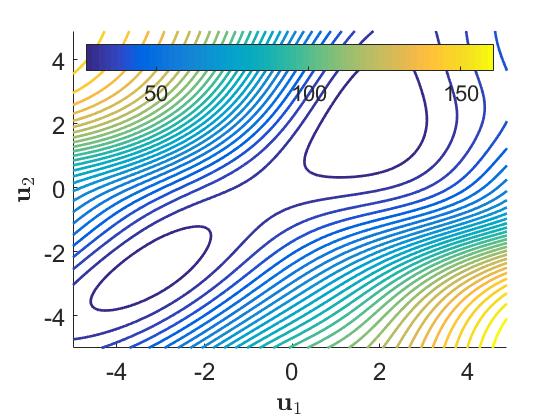}
		\caption{Tanh}
	\end{subfigure}%
	\begin{subfigure}{.33\textwidth}
		\centering 
		\includegraphics[width=0.99\linewidth]{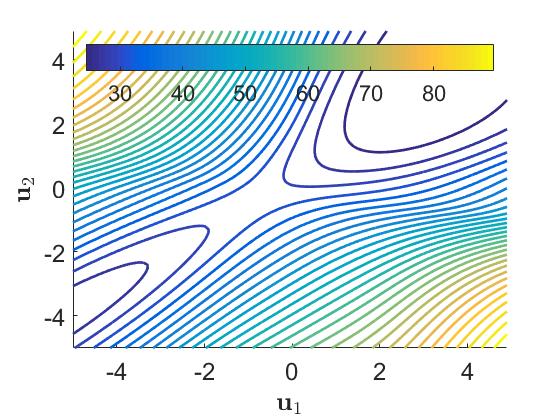}
		\caption{Softsign}
	\end{subfigure}%
	
	\begin{subfigure}{.33\textwidth}
		\centering 
		\includegraphics[width=0.99\linewidth]{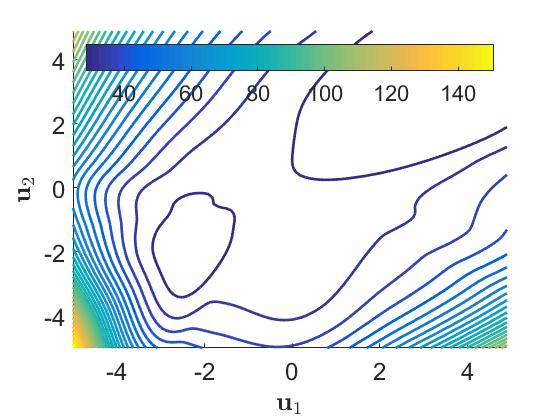}
		\caption{ReLU}
	\end{subfigure}%
	\begin{subfigure}{.33\textwidth}
		\centering
		\includegraphics[width=0.99\linewidth]{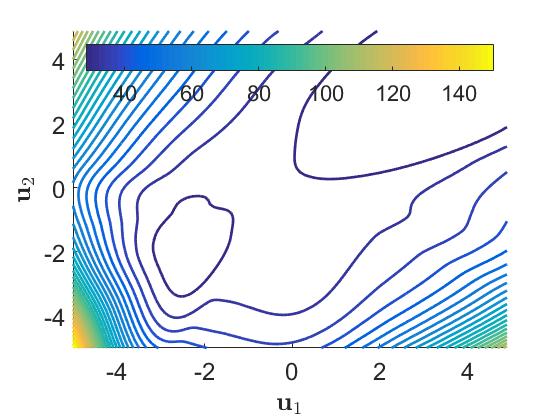}
		\caption{leaky ReLU}
	\end{subfigure}%
	\begin{subfigure}{.33\textwidth}
		\centering
		\includegraphics[width=0.99\linewidth]{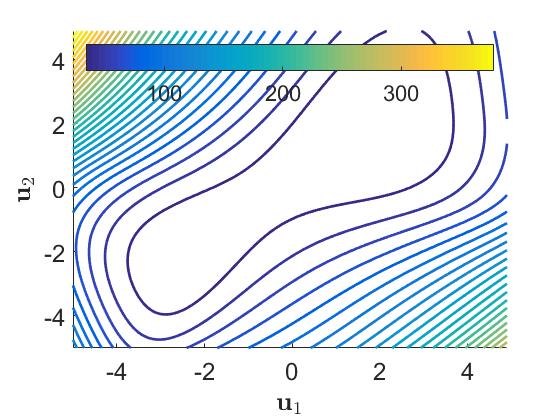}
		\caption{ELU}
	\end{subfigure}%
	
	\caption{Contours of the mean squared error loss function along two random perpendicular unit directions, $\boldsymbol{u}_1$ and $\boldsymbol{u}_2$, computed for a single hidden layer feedforward neural network using different activation functions. The hidden layer consists of 10 hidden nodes and the architecture is applied to the classic Iris dataset \citep{Fisher1936}.}
	\label{fig_conts}
\end{figure}

We consider the difference in loss function characteristics of the selected AFs for a simple neural network problem presented in Figure~\ref{fig_conts}. The contours of the mean squared error (MSE) loss function are depicted for a single hidden layer feedforward neural network with 10 hidden nodes, fitted to the Iris dataset \citep{Fisher1936}. The plots are generated by taking steps $\alpha_{\{1\}} \in [-5,-4.5,\dots,5] $ and $\alpha_{\{2\}} \in [-5,-4.5,\dots,5] $ along the two random perpendicular unit directions, $ \boldsymbol{u}_1$ and $ \boldsymbol{u}_2$, as inspired by \cite{Li2017}.

The contours of the Sigmoid AF represent a smooth loss function, containing a single minimum, where difference in function value is $\pm15$ over the sampled domain. The use of the Tanh AF results in a larger range in function value of over $\pm120$ and the emergence of an additional local minimum. Although the loss function range of Softsign is between that of Sigmoid and Tanh at $\pm 60$, in this case the two local minima drift further apart in the sampled domain. A characteristic feature of the saturation class of AFs is, that the loss function contours are smooth. Amongst the sparsity class of AFs, the ReLU AF retains the multi-modal nature of Tanh and Softsign, but demonstrates abrupt changes in contour characteristics. The modification of ReLU to include leaky gradients (leaky ReLU) softens these abrupt characteristics slightly, evidenced by the contours around the local minimum at $\alpha_{\{1\}} \approx -2$ and $\alpha_{\{2\}} \approx -2$. However, ELU impacts loss characteristics the most within the sparsity class, as it smooths out the contours of the loss and brings the two local minima closer together. ELU also results in a larger range of function value over the sampled domain, encompassing a change of $\pm 300$ compared to $\pm 120$ for ReLU and leaky ReLU respectively. 

It is clear, that the choice of AF can significantly influence loss function landscape characteristics. By extension, these changes translate to the respective loss function gradients. Previous studies have confirmed, that loss functions with higher curvature cause SNN-GPPs to be more localized in space \citep{Kafka2019b}. Consequently, the aim is to investigate and quantify the choice of AF with regards to the performance of GOLS in determining step sizes for dynamic MBSS neural network training; and if it is adversely affected, what can be done to improve or restore performance.

\section{Connections: Gradient-only line searches and activation functions}

\label{sec_connections}

Consider neural network loss functions formulated as: 
\begin{eqnarray}
\mathcal{L}(\boldsymbol{x}) = \frac{1}{M} \sum_{b=1}^{M} \ell (\hat{\boldsymbol{t}}_b(\boldsymbol{x});\;\boldsymbol{t}_b),
\label{eq:loss}
\end{eqnarray}
where $\boldsymbol{x}\in \mathcal{R}^p$ denotes the vector of weights parameterizing the neural network, the training dataset of $M$ samples is given by $\{\boldsymbol{t}_1,\dots,\boldsymbol{t}_M\}$, and $\ell(\hat{\boldsymbol{t}}_b(\boldsymbol{x});\;\boldsymbol{t}_b)$ is the elemental loss evaluating $\boldsymbol{x}$ (via neural network model prediction $ \hat{\boldsymbol{t}}_b(\boldsymbol{x})$) in terms of the training sample $\boldsymbol{t}_b$. Implementing Backpropagation \citep{Werbos1994} allows for efficient evaluation of the analytical gradient of $\mathcal{L}(\boldsymbol{x})$ with regards to $\boldsymbol{x}$: 
\begin{eqnarray}
\nabla\mathcal{L}(\boldsymbol{x}) = \frac{1}{M} \sum_{b=1}^{M} \nabla\ell (\hat{\boldsymbol{t}}_b(\boldsymbol{x});\;\boldsymbol{t}_b).
\label{eq:lossgrad}
\end{eqnarray}

When $\mathcal{L}(\boldsymbol{x})$ and $\nabla\mathcal{L}(\boldsymbol{x})$ are evaluated using the full training dataset of $M$ samples, the smoothness and continuity characteristics of both $\mathcal{L}(\boldsymbol{x})$ and $\nabla\mathcal{L}(\boldsymbol{x})$ are subject only to the smoothness and continuity characteristics of the AFs used in the neural network that constructs $\hat{\boldsymbol{t}}_b(\boldsymbol{x})$.


In order to conduct neural network training, the loss function in Equation (\ref{eq:loss}) is minimized. Consider the standard gradient-descent update: $ \boldsymbol{x}_{n+1} = \boldsymbol{x}_n - \alpha \nabla \mathcal{L}(\boldsymbol{x}_n) $\citep{Arora2011}. An iteration, $n$, is performed when the parameters, $\boldsymbol{x}_n$, are updated to a new state, $\boldsymbol{x}_{n+1}$. To determine step size, $\alpha$, line searches are performed at every iteration, $n$, of the training algorithm in pursuit of a good minimum. Thus an iteration, $n$, encompasses exactly one line search. Line searches are conducted by finding the optimum of a univariate function, $F_n(\alpha)$, constructed from current solution, $\boldsymbol{x}_n$, along a search direction, $\boldsymbol{d}_n $. If full-batch sampling is implemented in univariate function $F_n(\alpha)$, we define:
\begin{equation}
\mathcal{F}_n(\alpha) = f(\boldsymbol{x}_n(\alpha)) = \mathcal{L}(\boldsymbol{x}_n + \alpha \boldsymbol{d}_n),
\label{eq_linesearch}
\end{equation}
with corresponding directional derivative
\begin{equation}
\mathcal{F}'_n(\alpha) = \frac{d F_n(\alpha)}{d \alpha} = \boldsymbol{d}_n^T \cdot \nabla \mathcal{L}(\boldsymbol{x}_{n} + \alpha \boldsymbol{d}_n ).
\label{eq_lineg}
\end{equation}

\begin{figure}[h]
	\centering
	\includegraphics[width=0.9\linewidth]{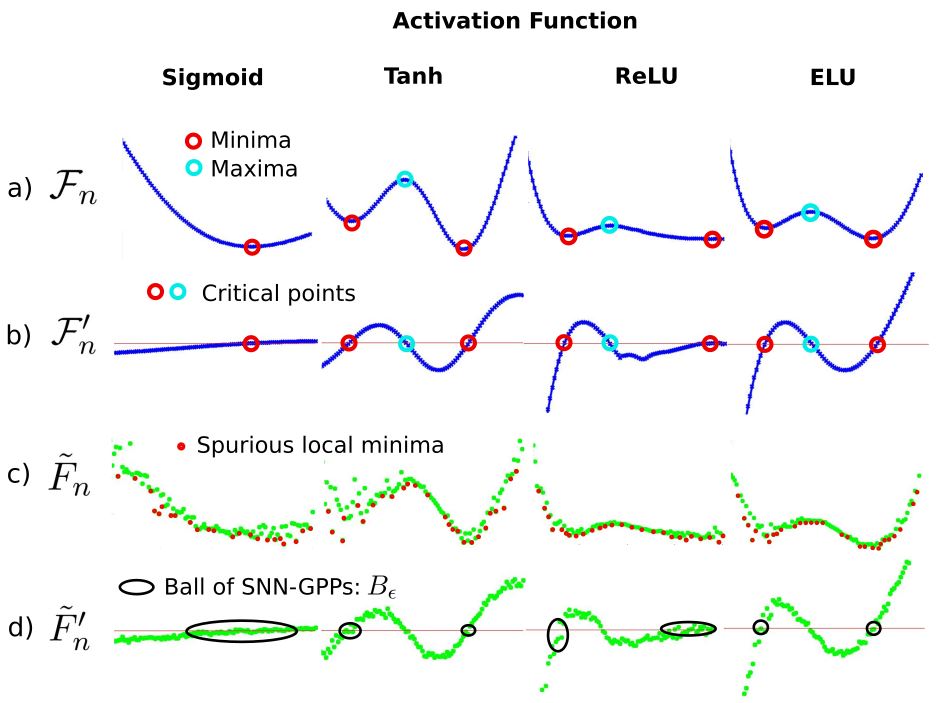}
	\caption{Univariate functions and directional derivatives of (a) $\mathcal{F}_n$ and (b) $\mathcal{F}'_n$, using full-batch sampling, and (c) $\tilde{F}_n$ and (d) $\tilde{F}'_n$, using dynamic mini-batch sub-sampling with a selection of activation functions.}
	\label{fig_LS_sum}
\end{figure}

Figures~\ref{fig_LS_sum}(a) and (b) show examples of $\mathcal{F}_n(\alpha)$ and corresponding directional derivative $\mathcal{F}'_n(\alpha)$ for different AFs. For compactness, the explicit dependency on variables such as $\alpha$ is dropped in further discussions, unless specifically required. In Figure~\ref{fig_LS_sum}, $\mathcal{F}_n$ and $\mathcal{F}'_n$ are constructed along the normalized search direction $\boldsymbol{d}_n = \frac{\boldsymbol{u}_1 + \boldsymbol{u}_2}{\| \boldsymbol{u}_1 + \boldsymbol{u}_2 \|_2}$ in our illustrative example introduced in Section~\ref{sec_intro}. Note, how all instances of $\mathcal{F}_n$ are continuous. This means that minimization line searches \citep{Arora2011} can be used to determine step sizes for training in full-batch sampled loss functions. The minimizer of $\mathcal{F}_n$, namely $\alpha^*$, subsequently becomes the step size for iteration $n$, i.e. $\alpha_{n,I_n}=\alpha^*$, where $I_n$ is the number of function evaluations required during the line search to find the optimum at iteration $n$. This notation can also be used to describe fixed step sizes. In such cases, $\alpha_{n,I_n}$ is a predetermined constant value over every iteration, and $I_n=1$.

However, modern datasets and corresponding network architectures have memory requirements that exceed current computational resources \citep{Krizhevsky2012}. Therefore, only a fraction of the training data, $\mathcal{B} \subset \{1,\dots,M\}$ with $|\mathcal{B}| \ll M$, is loaded into memory at a given time. This is referred to as mini-batch sub-sampling (MBSS). Omitting training data to construct mini-batches invariably results in a sampling error associated with the MBSS loss, compared to the full-batch sampled loss. Some training approaches employ static MBSS, where mini-batches are fixed for the minimum duration of a search direction, $\boldsymbol{d}_n$ \citep{Friedlander2011,Bollapragada2017,Kungurtsev2018,Kafka2019jogo}. Alternatively, dynamic MBSS can be implemented, where a new mini-batch is resampled for every evaluation, $i$, of the loss function. It has been shown that dynamic MBSS, also referred to as approximate optimization \citep{Bottou2010}, can benefit a given training algorithm by exposing it to larger amounts of information per search direction \citep{Kafka2019jogo}. We therefore define dynamic MBSS approximations of $\mathcal{L}(\boldsymbol{x})$ and $\nabla\mathcal{L}(\boldsymbol{x})$ respectively, as:

\begin{equation}
\tilde{L}(\boldsymbol{x}) = \frac{1}{|\mathcal{B}_{n,i}|} \sum_{b\in \mathcal{B}_{n,i}} \ell (\hat{\boldsymbol{t}}_b(\boldsymbol{x});\;\boldsymbol{t}_b),
\label{eq:lossgradbatch}
\end{equation}
and 
\begin{equation}
\tilde{\boldsymbol{g}}(\boldsymbol{x}) = \frac{1}{|\mathcal{B}_{n,i}|} \sum_{b\in \mathcal{B}_{n,i}} \nabla\ell (\hat{\boldsymbol{t}}_b(\boldsymbol{x});\;\boldsymbol{t}_b).
\label{eq:g_lossgradbatch}
\end{equation}


Note, that the mini-batch, $\mathcal{B}_{n,i}$, sampled for an instance of $\tilde{L}(\boldsymbol{x})$ and $\tilde{\boldsymbol{g}}(\boldsymbol{x})$ is consistent between the pair, while only at a new evaluation of pair $\tilde{L}(\boldsymbol{x})$ and $\tilde{\boldsymbol{g}}(\boldsymbol{x})$ is $\mathcal{B}_{n,i}$ resampled \citep{Kafka2019jogo}. However, the act of abruptly alternating between sampling errors associated different mini-batches $\mathcal{B}_{n,i}$, interrupts the smoothness and continuity characteristics of the loss function, irrespective of the choice of AF used in the neural network architecture. This results in point-wise discontinuous loss, $\tilde{L}(\boldsymbol{x})$, and gradient, $\tilde{\boldsymbol{g}}(\boldsymbol{x})$, functions. Although $\E [ \tilde{L}(\boldsymbol{x}) ] = \mathcal{L}(\boldsymbol{x})$ and $\E [ \tilde{\boldsymbol{g}}(\boldsymbol{x}) ] = \nabla\mathcal{L}(\boldsymbol{x})$ \citep{Tong2005}, the probability of encountering critical points, $\tilde{g}(\boldsymbol{x}^*)=\bar{\boldsymbol{0}}$, is infeasibly low in dynamic MBSS loss functions. Additionally, the point-wise discontinuities between consecutive evaluations of $\tilde{L}(\boldsymbol{x})$ result in the emergence of spurious candidate local minima \citep{Wilson2003,Schraudolph2003,Schraudolph2006}, which have been shown to be approximately uniformly distributed over the loss landscape \citep{Kafka2019b}.

{
	By substituting $\tilde{L}(\boldsymbol{x})$ and $\tilde{\boldsymbol{g}}(\boldsymbol{x})$ into Equations (\ref{eq_linesearch}) and (\ref{eq_lineg}) respectively, we obtain dynamic MBSS univariate function $\tilde{F}_n(\alpha)$ and corresponding directional derivative $\tilde{F}'_n(\alpha)$. Consider Figures~\ref{fig_LS_sum}(c) and (d) for a range of AFs. Note, how sampling $\mathcal{B}_{n,i}$ uniformly from the full training dataset in Figure~\ref{fig_LS_sum}(c), results in local minima all along the search direction. Although directional derivatives Figure~\ref{fig_LS_sum}(d) get close to zero, none are a critical point, i.e. $\tilde{F}'_n(\alpha^*) \neq 0$. This is best illustrated by the first local minimum along the search direction for ReLU. Using full-batch sampling, a clear local minimum, $\mathcal{F}_n(\alpha^*)$, can be observed in Figure~\ref{fig_LS_sum}(a)(ReLU) with  corresponding critical point, $\mathcal{F}'_n(\alpha^*)= 0$, in Figure~\ref{fig_LS_sum}(b)(ReLU). Both are indicated by the first red circle from left to right along $\mathcal{F}_n$ and $\mathcal{F}'_n$ respectively. Using dynamic MBSS in Figures~\ref{fig_LS_sum}(c)(ReLU) and (d)(ReLU), none of the directional derivatives are critical points, $\tilde{F}'_n \neq 0$, and local minima are located all along $\tilde{F}_n$, illustrated by small red points. 
}

The discontinuities in $\tilde{F}_n$ make minimization ineffective in determining step sizes in dynamic MBSS loss functions \citep{Kafka2019jogo}. Historically, this has led to the popularity of subgradient methods for neural network training, using {\it a priori} determined step sizes.\citep{Schraudolph1999,Boyd2003,Smith2015}. Line searches were first introduced into dynamic MBSS loss functions by \cite{Mahsereci2017a}, determining step sizes by using probabilistic surrogates along $\tilde{F}_n$ to estimate the location of optima. An alternative approach is the use of Gradient-Only Line Searches (GOLS) \citep{Kafka2019jogo,Kafka2019}, which employ an extension of the gradient-only optimality criterion \citep{Wilke2013,Snyman2018}, namely the {\it Stochastic Non-Negative Associative Gradient Projection Point} (SNN-GPP) \citep{Kafka2019jogo}, given as follows:

\begin{definition}{SNN-GPP:}
	A stochastic non-negative associated gradient projection point (SNN-GPP) is defined as any point, $\boldsymbol{x}_{snngpp}$, for which there exists 	$r_u > 0$ such that 
	\begin{equation}
	\nabla f(\boldsymbol{x}_{snngpp} + \lambda \boldsymbol{u} )\boldsymbol{u} \geq 0,\;\;  \forall\; \boldsymbol{u} \in \left\{\boldsymbol{y}\in\mathbb{R}^p \;|\; \| \boldsymbol{y} \|_2 = 1\right\},\;\;\forall \;\lambda \in (0,r_u],
	\end{equation}	
	with non-zero probability. \citep{Kafka2019jogo}
	\label{def_snngpp}
\end{definition}

{
	Subsequently, a ball, $B_\epsilon$, exists that bounds all possible SNN-GPPs of a surrounding neighbourhood, where each neighbourhood contains one true optimum:
	\begin{definition}{ $B_\epsilon$:}
		Consider a dynamic mini-batch sub-sampled loss function $\tilde{L}(\boldsymbol{x})$, of a continuous, smooth and convex full-batch loss function $\mathcal{L}(\boldsymbol{x})$, such that each sampled mini-batch with associated $L(\boldsymbol{x})$ that is used to evaluate $\tilde{L}(\boldsymbol{x})$ has the same smoothness, continuity and convexity characteristics as $\mathcal{L}(\boldsymbol{x})$. Then there exists a ball, 
		\begin{equation}
		B_\epsilon(\boldsymbol{x}) = \{\boldsymbol{x} \in \mathbb{R}^p : \|\boldsymbol{x}-\boldsymbol{x}^*\|_2 <\epsilon,\;0<\epsilon<\infty,\;\boldsymbol{x}\neq \boldsymbol{x}^* \},
		\label{eq_ball}
		\end{equation} 
		that contains all the stochastic non-negative gradient projection points (SNN-GPPs), where $\boldsymbol{x}^*$ is the minimum of $\mathcal{L}(\boldsymbol{x})$. \citep{Kafka2019jogo}
		\label{thm:ball}
	\end{definition}
}

Along any univariate function, $F_n$, an SNN-GPP manifests as a sign change in the directional derivative from negative, $F'_n<0$, to positive, $F'_n>0$, along the descent direction. In the deterministic, full-batch setting, $\mathcal{F}_n$, the SNN-GPP reduces to the critical point associated with a local minimum, i.e. $\mathcal{F}_n(\alpha_{snngpp}) = \mathcal{F}_n(\alpha^*)$, and ball $B_\epsilon$ reduces to a single point. In the stochastic setting of dynamic MBSS losses, ball $B_\epsilon$ has a finite range that is dependent on the variance of the directional derivative as well as the expected curvature $\frac{\delta E[\tilde{F}'_n]}{\delta \alpha}$ in a neighbourhood \citep{Kafka2019b}. 

The SNN-GPP and $B_\epsilon$ can be visually illustrated in Figure~\ref{fig_LS_sum}(d). With the Sigmoid activation function, there is a single neighbourhood, with a large ball $B_\epsilon$ containing all SNN-GPPs. There exist numerous sign changes from negative to positive along the descent direction in $B_\epsilon$, due to the variance of $\tilde{F}'_n$ and the slow change in $\tilde{F}'_n$ along $\alpha$. In the case of Tanh, ReLU and ELU, there are two neighbourhoods in which SNN-GPPs can be found. These neighbourhoods are separated by a maximum, as demonstrated by $\mathcal{F}'_n$. Note, how the SNN-GPP definition ignores maxima, as it considers only sign changes from negative to positive along the descent direction. The $\tilde{F}'_n$ plots of Tanh, ReLU and ELU demonstrate how the size of ball $B_\epsilon$ decreases, with a decrease in variance and an increase in expected curvature. These $\tilde{F}'_n$ plots also show, that the characteristics of $B_\epsilon$ can change in different neighbourhoods of the loss function, and vary according to each AF.

It has been shown, that an exact GOLS will converge to an SNN-GPP within ball $B_\epsilon$ \citep{Kafka2019jogo}. Therefore, GOLS determine the step size at iteration $n$ of a training algorithm, by locating an SNN-GPP such that $\alpha_{n,I_n}=\alpha_{snngpp}$. It has also been demonstrated, that the Gradient-Only Line Search that is Inexact (GOLS-I) behaves in a manner consistent with Lyapunov's global stability theorem \citep{Lyapunov1992,Kafka2019}. The latter proof was developed in the context of loss functions that are positive, coercive and AFs that result in strict descent \citep{Kafka2019}. Subsequently, this paper explores how GOLS perform with a larger range of AFs; and consider the implications a given AF may have on a neural network architecture for a given problem.

\section{Contribution}

In this paper we empirically study the interaction between activation functions and neural network architectures, when using Gradient-Only Line Searches (GOLS) to determine step sizes for dynamic MBSS loss functions. In our investigations we consider six activation functions, as introduced in Section~\ref{sec_intro}, in the context of 1) shallow and deep feedforward classification networks, and 2) architectural features such as batch normalization and skip connections. To this end, we use 13 datasets to construct a range of training problems, where we primarily use the Gradient-Only Line Search that is Inexact (GOLS-I) \citep{Kafka2019jogo} to determine step sizes for training. Depending on the nature of the investigation, the Gradient-Only Line Search with Bisection (GOLS-B) \citep{Kafka2019jogo} and fixed step sizes are sporadically used to be benchmarks against which the performance of GOLS-I can be compared. 
Overall, our investigations demonstrate that GOLS-I is effective in determining step sizes in a range of feedforward neural network architectures using different activation functions. However, we also give examples, where activation function selection can significantly impede training performance with GOLS-I. We show that these difficulties can be alleviated by modifying the network architecture of a given problem. Therefore, this paper serves as a practical guide for neural network practitioners to improve the construction of network architectures that promote efficient training using GOLS.



\section{Empirical Study}

In our studies we consider three different types of training problems, namely:
\begin{enumerate}
	\item Foundational problems: Using small classification datasets with various feedforward neural network architectures.
	\item MNIST with NetII: A training problem used by \cite{Mahsereci2017a} to explore early stochastic line searches.
	\item CIFAR10 with ResNet18: A state of the art architecture including skip-connections \citep{He2016} and batch normalization \citep{Ioffe2015}.
\end{enumerate}

The foundational training problems are used to explore the influence of AFs on training performance in the context of 1) hidden layer height and depth, 2) GOLS-I and GOLS-B, as well as constant step sizes; and 3) full-batch versus dynamic MBSS training. The NetII problem is used to demonstrate the potential sensitivity of training problems to the choice of AF, and subsequently show the corrective effect of batch normalization on training.
Skip connections are another architectural consideration of interest with different AFs in the context of GOLS. The relationship between AFs and neural networks with skip connections is explored using the ResNet18 architecture with the CIFAR10 dataset, as adapted from the implementation by \cite{Kuangliu2018}. 

The datasets used in this study, along with and their respective properties, are listed in Table~\ref{tbl_datasets}. All datasets were scaled using the standard transform (Z-score). For the foundational problems (spanning datasets 1 to 11), the standard transform was applied for each individual input D, while for MNIST and CIFAR10 the standard transform was applied over each image channel. For MNIST, there is a single grey scale channel of 28x28 pixels (total of 784 inputs), while CIFAR10 has 3 colour channels of 32x32 pixels each (total of 3072 inputs). Since the small datasets are not separated into training and test datasets by default, we divided them manually into training, validation and test datasets with a ratio of 2:1:1 respectively. We choose this division to demonstrate that the manual construction of validation and test datasets resulted in representative, unbiased hold-out datasets. Therefore, we expect similar performance between validation and test datasets. Conversely, both MNIST and CIFAR10 datasets have been predetermined test datasets, which are subsequently used.

\begin{table}[h!]
	\centering
	\scalebox{0.65}{
		\begin{tabular}{|c|l|l|c|c|c|}
			\hline  
			\textbf{No.} & \textbf{Dataset name} & \textbf{Author} & \textbf{Observations}  & \textbf{Inputs}, $D$ & \textbf{Classes}, $K$ \\ 
			\hline 1 & Iris & \cite{Fisher1936} & 150 & 4 & 3 \\ 
			\hline 2 & Glass1 & \cite{Prechelt1994} & 214 & 9 & 6 \\ 
			\hline 3 & Horse1 & \cite{Prechelt1994} & 364 & 58 & 3 \\ 
			\hline 4 & Forests & \cite{Johnson2012} & 523 & 27 & 4\\ 
			\hline 5 & Simulation failures & \cite{Lucas2013} & 540 & 20 & 2\\ 
			\hline 6 & Soybean1 & \cite{Prechelt1994} & 683 & 35 & 19 \\ 
			\hline 7 & Card1 & \cite{Prechelt1994} & 690  & 51 & 2  \\ 
			\hline 8 & Cancer1 & \cite{Prechelt1994} & 699  & 9 & 2  \\ 
			\hline 9 & Diabetes1 & \cite{Prechelt1994} & 768  & 8 & 2 \\ 
			\hline 10 & Heartc1 & \cite{Prechelt1994} & 920 & 35 & 2 \\ 
			\hline 11 & Biodegradable compounds & \cite{Mansouri2013} & 1054 & 41 & 2 \\ 
			\hline 12 & MNIST & \cite{Lecun1998} & 70000 & 784 & 10\\ 
			\hline 13 & CIFAR10 & \cite{Krizhevsky2009} & 60000 & 3072 & 10 \\ 
			\hline 
	\end{tabular}}
	\caption{Properties of the datasets considered for the this study.}
	\label{tbl_datasets} 
\end{table}

Table~\ref{tbl_nets} summarizes the 11 investigations performed in this study on the corresponding neural network training problems. For the foundational problems we implement shallow nets with the number of hidden nodes being half of the input dimensions, $\frac{D}{2}$ and twice that of the input dimensions, $2D$. We also implement a deep architecture with 6 hidden layers of $2D$ nodes. We conduct training limited by iterations for all training problems except NetII, which is limited in the number of function evaluations, as prescribed by \cite{Mahsereci2017a}. We couple each of the training problems with the activation functions discussed in Section~\ref{sec_connections}. Constructed dynamic MBSS, search directions $\boldsymbol{d}_n = -\tilde{\boldsymbol{g}}(\boldsymbol{x})$ were supplied by the line search stochastic gradient descent (LS-SGD) algorithm \citep{Robbins1951,Kafka2019jogo} for all training runs accept those of the deep architecture applied to the foundational problems, which uses Adagrad \citep{Duchi2011,Kafka2019jogo} instead. We adopt the convention whereby the name of a method is the combination of the line search used to determine the step size, and the algorithm used to provide the search direction. For example, using GOLS-I to determine step sizes for Adagrad is denoted "GOLS-I Adagrad". As indicated in Table~\ref{tbl_nets}, step sizes for LS-SGD were predominantly determined using GOLS-I, while alternatively GOLS-B and fixed step sizes were implemented, depending on the investigation performed. The fixed step size, $\alpha_{n,I_n}=0.05$, used in investigation 4 was manually tuned to give good training performance for the foundational problems with the ReLU AF. The fixed step sizes chosen for investigation 8 were selected in order to highlight a range of training performance, from slow to unstable, for the NetII training problem with the ReLU AF.

\begin{table}[h]
	\centering
	\scalebox{0.68}{
		\begin{tabularx}{\textwidth}{|p{7mm}|p{11mm}|p{27mm}|p{6mm}|p{24mm}|p{11mm}|p{8mm}|X|}
			\hline 
			\textbf{Inv. \#} & \textbf{Data- set \#} & \textbf{Hidden layer structure} & \textbf{BN} & \textbf{Step size method} & \textbf{Train limit} & \textbf{$|\mathcal{B}_{n,i}|$} & \textbf{Loss} \\ 
			\hline 
			\textbf{1} & 1-11 & $[\frac{D}{2}]$ & No & GOLS-I SGD & 3000 Its. & 32 & MSE \\ 
			\hline 
			\textbf{2} & 1-11 & $[2D]$ & No & GOLS-I SGD & 3000 Its. & 32 & MSE \\ 
			\hline 
			\textbf{3} & 1-11 & $[2D,2D,2D,$ $2D,2D,2D]$ & No & GOLS-I Adagrad & 3000 Its. & 32 & MSE \\ 
			\hline 
			\textbf{4} & 1-11 & $[2D]$ & No &  $\alpha_{n,I_n}=0.05$ with LS-SGD & 3000 Its. & 32 & MSE \\ 
			\hline 
			\textbf{5} & 1-4 & $[2D]$ & No & GOLS-B SGD & 3000 Its. & 64 & MSE \\ 
			\hline 
			\textbf{6} & 1-4 & $[2D]$ & No & GOLS-B SGD & 3000 Its. & $M$ & MSE \\ 
			\hline 
			\textbf{7} & 12 & $[1000,500,250]$ & No & GOLS-I SGD & 40,000 FEs & 100 & MSE \\ 
			\hline 
			\textbf{8} & 12 & $[1000,500,250]$ & No & $\alpha_{n,I_n}=0.1$, $\alpha_{n,I_n}=0.01$,  $\alpha_{n,I_n}=0.001$ with LS-SGD & 40,000 FEs & 100 & MSE \\ 
			\hline 
			\textbf{9} & 12 & $[1000,500,250]$ & Yes & GOLS-I SGD & 40,000 FEs & 100 & MSE \\ 
			\hline 
			\textbf{10} & 13 & ResNet18 & Yes & GOLS-I SGD & 40,000 Its. & 128 & BCE \\ 
			\hline 
			\textbf{11} & 13 & ResNet18 & No & GOLS-I SGD & 40,000 Its. & 128 & BCE \\ 
			\hline 
		\end{tabularx} 
	}
	\caption{Parameters and settings governing the implemented network architectures (with and without batch normalization (BN)) regarding their training in various investigations (Inv.).}
	\label{tbl_nets}	
\end{table}

All training runs were conducted using PyTorch 1.0 \citep{PyTorch1}. By default, He initialization \citep{He2015} is used for networks implementing ReLU and leaky ReLU AFs, while Xavier initialization \citep{Glorot2010} is used for networks with the remaining AFs considered in this study. For the foundational and NetII training problems, 10 training runs were conducted for each dataset and AF combination, whereas for CIFAR10 with ResNet18, a single training run per AF is performed.

\subsection{Software}
In pursuit of transparency and reproducibility, we have made code available at \url{https://github.com/gorglab/GOLS}. The repositories posted include user-friendly versions of the source code used in our investigations. These include GPU compatible examples of all the GOLS methods developed in \cite{Kafka2019jogo}, including GOLS-I and GOLS-B as used in this study. Example codes are self contained and accessible, such as to create an environment suited to exploring the properties of GOLS.

\section{Results}

The results of our empirical study are ordered according to the training problems considered, namely 1) foundational problems, 2) the MNIST dataset with the NetII architecture, and 3) the CIFAR10 dataset with the ResNet18 architecture. Note, that the loss is used to evaluate training performance for the foundational problems, while the classification errors of the datasets are plotted to evaluate NetII and ResNet18. Note, that results given in terms of iterations are not representative of computational cost, as a number of function evaluations can be performed per iteration when line searches are implemented. However, giving results in terms of iteration allows for comparison between line searches with different costs, while assessing the reduction in loss (i.e. the quality) provided by the line searches.

\subsection{Foundational problems}

\label{sec_foundationalProbs}

\begin{figure}[h!]
	\centering
	\begin{subfigure}{.249\textwidth}
		\centering 
		\includegraphics[width=0.99\linewidth]{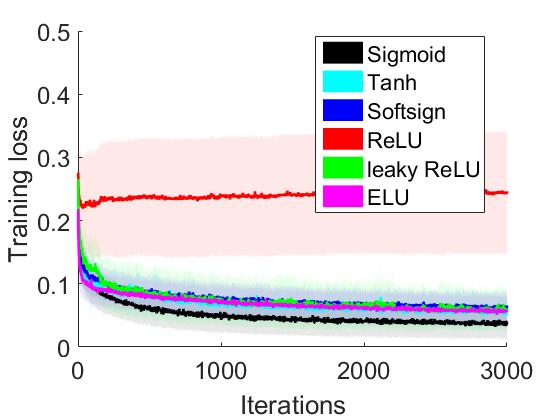}
	\end{subfigure}%
	\begin{subfigure}{.249\textwidth}
		\centering
		\includegraphics[width=0.99\linewidth]{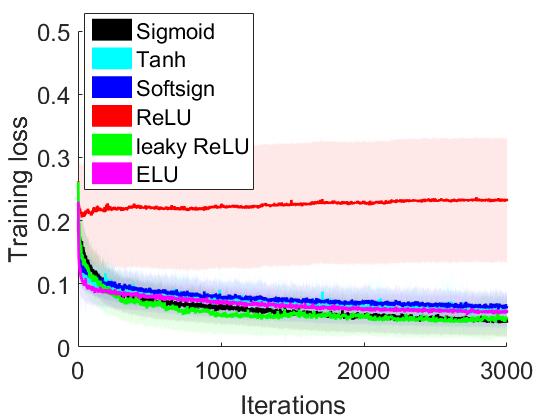}
	\end{subfigure}%
	\begin{subfigure}{.249\textwidth}
		\centering
		\includegraphics[width=0.99\linewidth]{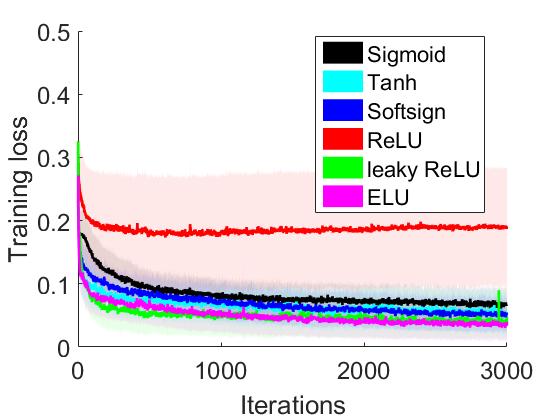}
	\end{subfigure}%
	\begin{subfigure}{.249\textwidth}
		\centering
		\includegraphics[width=0.99\linewidth]{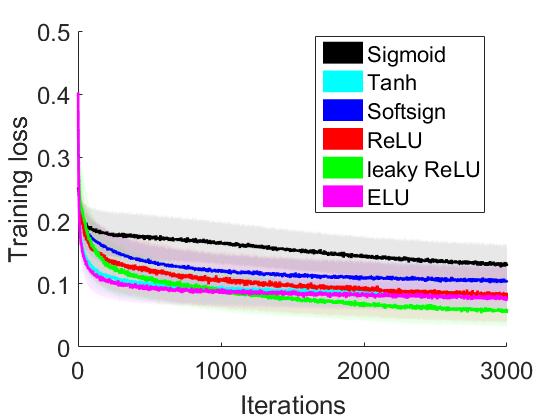}
	\end{subfigure}%

	\begin{subfigure}{.249\textwidth}
		\centering 
		\includegraphics[width=0.99\linewidth]{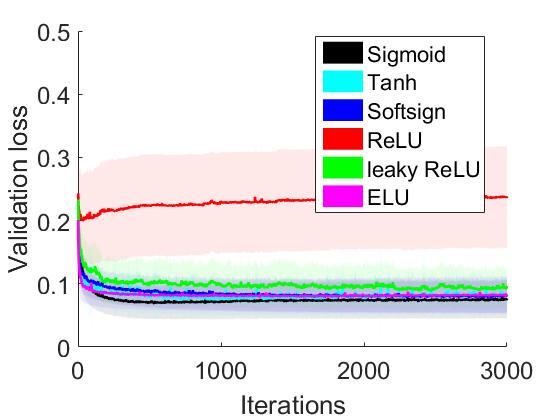}
	\end{subfigure}%
	\begin{subfigure}{.249\textwidth}
		\centering
		\includegraphics[width=0.99\linewidth]{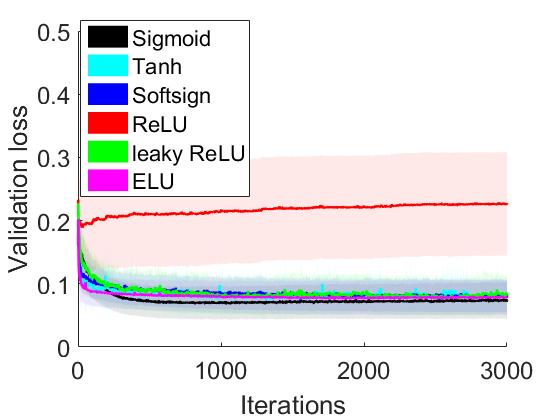}
	\end{subfigure}%
	\begin{subfigure}{.249\textwidth}
		\centering
		\includegraphics[width=0.99\linewidth]{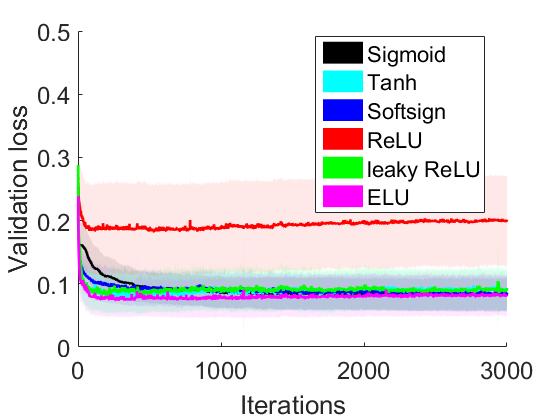}
	\end{subfigure}%
	\begin{subfigure}{.249\textwidth}
		\centering
		\includegraphics[width=0.99\linewidth]{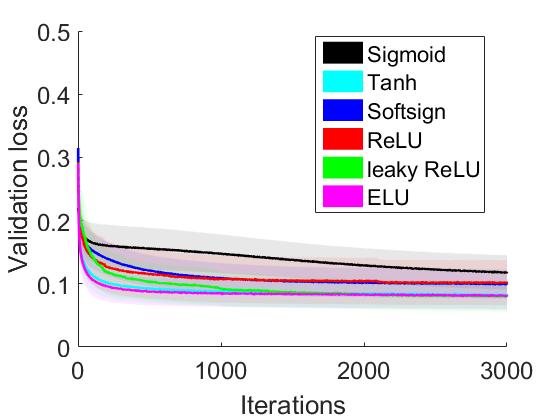}
	\end{subfigure}%
	
	\begin{subfigure}{.249\textwidth}
		\centering 
		\includegraphics[width=0.99\linewidth]{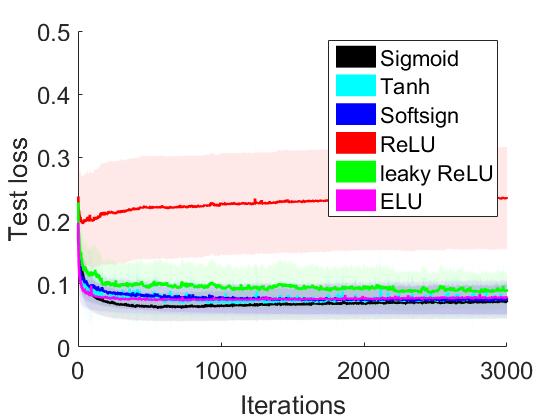}
	\end{subfigure}%
	\begin{subfigure}{.249\textwidth}
		\centering
		\includegraphics[width=0.99\linewidth]{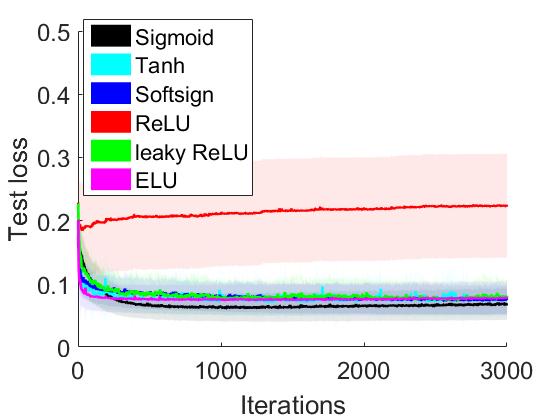}
	\end{subfigure}%
	\begin{subfigure}{.249\textwidth}
		\centering
		\includegraphics[width=0.99\linewidth]{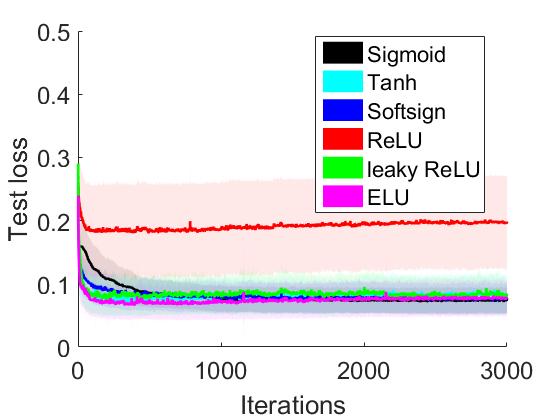}
	\end{subfigure}%
	\begin{subfigure}{.249\textwidth}
		\centering
		\includegraphics[width=0.99\linewidth]{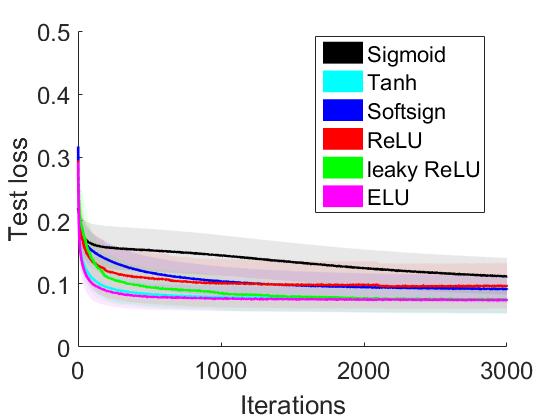}
	\end{subfigure}%
	
	\begin{subfigure}{.249\textwidth}
		\centering 
		\includegraphics[width=0.99\linewidth]{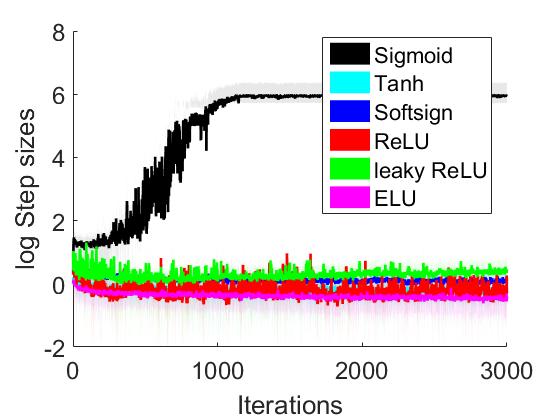}
		\caption{$\frac{D}{2}$, GOLS-I SGD}
	\end{subfigure}%
	\begin{subfigure}{.249\textwidth}
		\centering
		\includegraphics[width=0.99\linewidth]{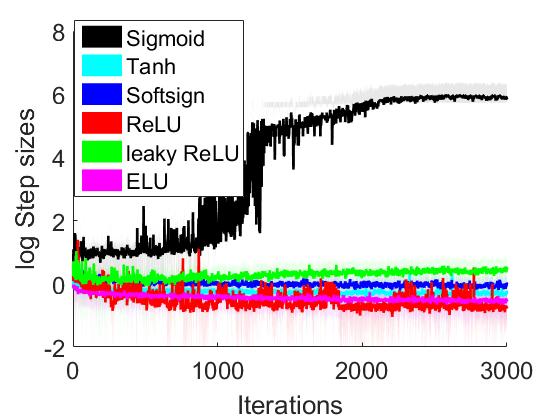}
		\caption{$2D$, GOLS-I SGD}
	\end{subfigure}%
	\begin{subfigure}{.249\textwidth}
		\centering
		\includegraphics[width=0.99\linewidth]{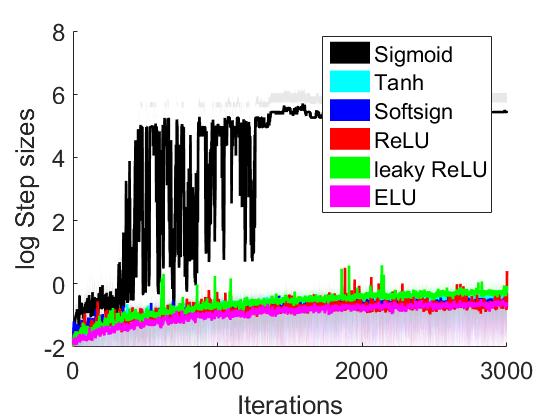}
		\caption{$2D(6)$, GOLS-I Ada.}
	\end{subfigure}%
	\begin{subfigure}{.249\textwidth}
		\centering
		\includegraphics[width=0.99\linewidth]{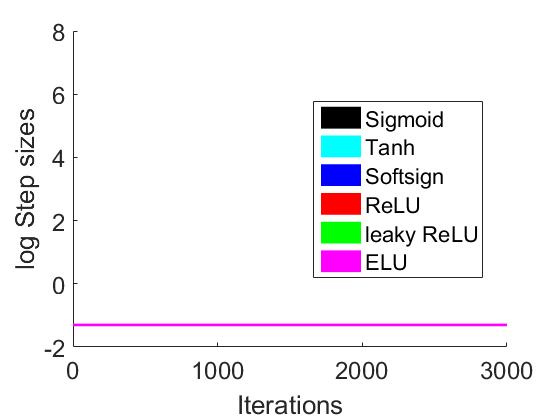}
		\caption{$2D$, $\alpha_{n,I_n}=0.05$}
	\end{subfigure}%
	
	\caption{Training, validation and test losses with corresponding log of step sizes for the foundational problems with various networks architectures, including (a) investigation 1: single hidden layer (HL) networks with $\frac{D}{2}$ hidden nodes, (b) investigation 2: single HL networks with $2D$ nodes and (c) investigation 3: networks with six HLs of $2D$ nodes using GOLS-I Adagrad. For (a) to (c), GOLS-I was used to determine step sizes, while (d) investigation 4 implements LS-SGD with fixed step sizes $\alpha_{n,I_n}=0.05$ for a single HL network with $2D$ nodes.}
	\label{fig_foundationals_mb}
\end{figure}

The average training, validation and test losses with corresponding average step sizes for the foundational problems are given in Figure~\ref{fig_foundationals_mb}. The results of the foundational problems are averaged over all the respective datasets and their corresponding 10 training runs. This allows a representative trend to be demonstrated for each AF over a number of datasets. The results of the first analysis, with hidden layers of size $\frac{D}{2}$, are shown in the first column, Figure~\ref{fig_foundationals_mb}(a). Firstly, it is evident that step sizes estimated by GOLS-I result in effective training over a range of datasets and AFs. The mean training loss continually decreases, while that of both the validation and test datasets increases after 500 iterations, indicating the onset of overfitting. The consistency between validation and test losses suggests, that both validation and test datasets are large enough to give unbiased assessment of the neural networks' generalization performance.

In investigation 1, shown in Figure~\ref{fig_foundationals_mb}(a), ReLU is convincingly the worst performer. Conversely, the Sigmoid AF is the best performer in training, while the performance between the remaining AFs is almost indistinguishable. The best validation and test losses also belong to Sigmoid, with a marginal advantage over the remaining AFs (accept ReLU, where the difference is significant). Interestingly, the step sizes of Sigmoid, presented in the last row of Figure~\ref{fig_foundationals_mb}(a), diverge significantly from those of the remaining AFs. We postulate that this phenomenon compensates for the Sigmoid's small derivative magnitudes, see Figure~\ref{fig_acts}(b). As mentioned, the maximum derivative magnitude of the Sigmoid AF is $0.25$, while for the remaining AFs it is 1. Cumulatively, this results in progressively smaller gradients for Sigmoid, as the training algorithm progresses closer towards an optimum. Subsequently, this prompts GOLS-I to increase the step sizes in the search for univariate SNN-GPPs. This is not common among the foundational problem datasets, but the larger step sizes of individual problems dominate the calculation of average step sizes.

The ReLU AF was introduced to promote sparsity within a neural network, which is meant to approximate brain processes observed in neuroscience \citep{Glorot2011}, where only a fraction of the network is used at a given time. However, ReLU was developed predominantly for large neural networks, which makes it unclear whether the bad performance observed in Figure~\ref{fig_foundationals_mb}(a) is due to the use of GOLS-I for determining learning rates, or due to the network architecture selected being too small. 

We therefore perform training runs with increased hidden layer sizes of $2D$ for investigation 2 shown in Figure~\ref{fig_foundationals_mb}(b). Hence, the hidden layers of network architectures trained in investigation 2 are 4 times larger than those considered in investigation 1. However, this increase has little effect on improving the training performance of ReLU. Instead, there is a shift between the relative performances of the remaining AFs. Subsequently, leaky ReLU outperforms the Sigmoid AF in training. However, this does not translate to generalization, where the Sigmoid still outperforms leaky ReLU. 


Presuming that the investigated architectures are still too small to cater for the sparsity that ReLU induces, we increase the number of hidden layers to 6 with $2D$ nodes each in investigation 3. To aid training with deeper layers, we employ the GOLS-I Adagrad for this analysis. The results shown in Figure~\ref{fig_foundationals_mb}(c) exhibit an average loss improvement of $0.0457$ for ReLU over the single hidden layer training runs performed for investigation 2 in Figure~\ref{fig_foundationals_mb}(b). It is notable, that the performance rankings between the remaining activation functions has again changed. In this case, the other AFs of the sparsity class (leaky ReLU and ELU), begin to dominate both training and generalization over AFs of the saturation class. The increasing step sizes in Figure~\ref{fig_foundationals_mb}(c)(last row) occur as GOLS-I corrects for the diminishing norm of Adagrad's search directions \citep{Duchi2011}, which is expected \citep{Kafka2019jogo}. Although GOLS-I determines useful learning rates that result in effective training for a range of AFs for a significantly larger network architecture, this analysis has failed to demonstrate a significant improvement in the training performance of ReLU with GOLS-I. This suggests, that neither architecture nor the search direction are dominant factors in explaining ReLU's performance in our experiments thus far. Therefore, the use of GOLS-I to determine the step sizes for ReLU architectures comes under closer scrutiny.

We determine GOLS-I's contribution to poor ReLU performance, by process of elimination. In investigation 4, shown in Figure~\ref{fig_foundationals_mb}(d), we substitute GOLS-I for the use of a manually tuned fixed step size of $\alpha_{n,I_n}=0.05$ with LS-SGD for all AFs. It is clear, that the training performance with ReLU improves significantly with $\alpha_{n,I_n}=0.05$. However, this is not coincidental, as the fixed step size was manually tuned for this purpose. This indicates, that the step sizes determined by GOLS-I are not effective for ReLU with the foundational training problems. Interestingly, the variance between training performances of the remaining AFs is higher with LS-SGD using fixed step size, than when implementing GOLS-I SGD. The Sigmoid AF performs significantly worse, due to its comparatively smaller derivatives, as discussed above. This confirms, that GOLS-I is capable of adapting its step sizes to the properties of different AFs in feedforward network architectures, with the exception of ReLU.


\subsection{A closer look at ReLU loss functions using GOLS-B SGD}

What makes ReLU the outlier among the considered AFs, is that it enforces sparsity in an absolute manner, i.e. the derivative in the negative input domain is exactly zero. Previous studies have shown, that the use of ReLU with the MSE loss can cause $\tilde{L}(\boldsymbol{x})$ and $\tilde{\boldsymbol{g}}(\boldsymbol{x})$ to be zero over a range of $\boldsymbol{x}$ \citep{Kafka2019b}. This breaks the assumptions of positivity, coerciveness and strict descent, namely those of Lyapunov's global stability theorem, which govern the convergence of GOLS \citep{Kafka2019}. The reason for GOLS-I's inability to train feedforward networks with ReLU is as follows: Conducting updates with step sizes that are too large, as is possible in an inexact line search such as GOLS-I, can cause numerous nodes within a ReLU network to enter the negative input domain. If the step sizes are large enough, the affected nodes remain "off" irrespective of the variance in the incoming data. Subsequently, large parts of the network may be "off" permanently, causing the gradient vector to have numerous zero-valued partial derivatives, i.e. become sparse. This results in no change to weights with zero-partial-derivatives during update steps, effectively terminating training for the deactivated portion of the architecture.

\begin{figure}[h!]
	\centering
	\begin{subfigure}{.249\textwidth}
		\centering 
		\includegraphics[width=0.99\linewidth]{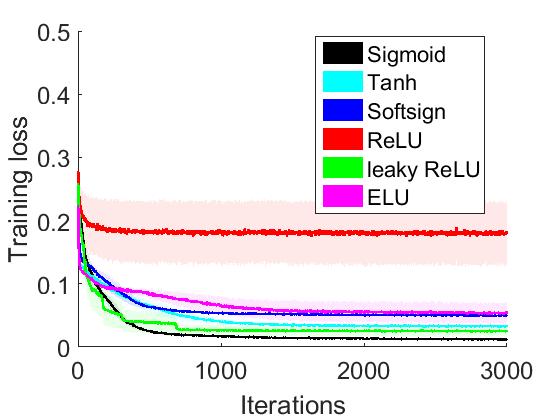}
	\end{subfigure}%
	\begin{subfigure}{.249\textwidth}
		\centering 
		\includegraphics[width=0.99\linewidth]{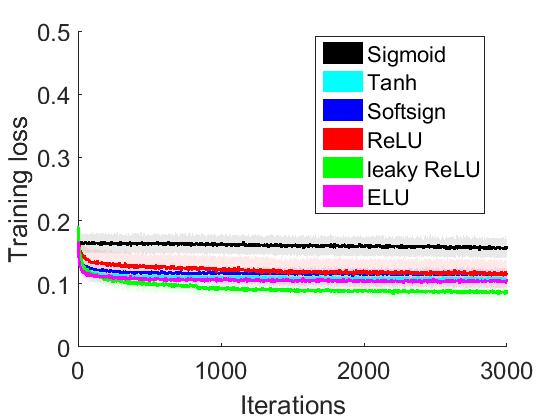}
	\end{subfigure}%
	\begin{subfigure}{.249\textwidth}
		\centering 
		\includegraphics[width=0.99\linewidth]{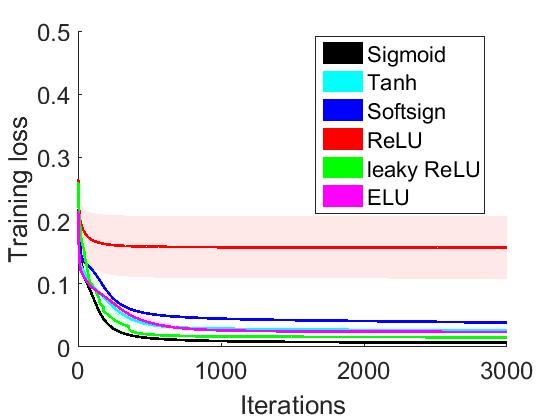}
	\end{subfigure}%
	\begin{subfigure}{.249\textwidth}
		\centering 
		\includegraphics[width=0.99\linewidth]{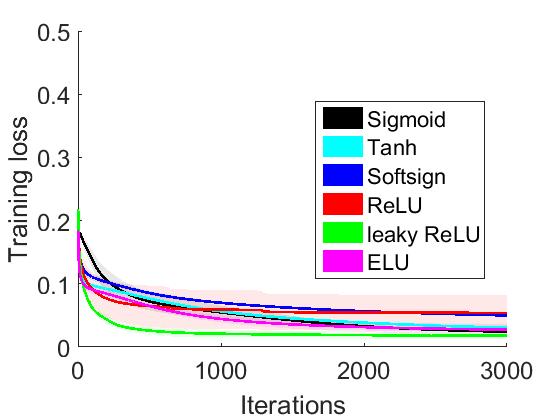}
	\end{subfigure}%
	
	\begin{subfigure}{.249\textwidth}
		\centering
		\includegraphics[width=0.99\linewidth]{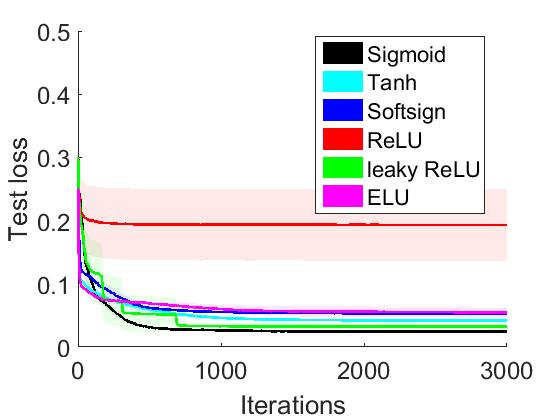}
	\end{subfigure}%
	\begin{subfigure}{.249\textwidth}
		\centering
		\includegraphics[width=0.99\linewidth]{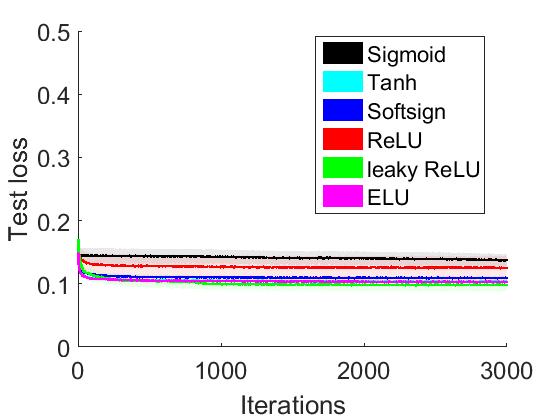}
	\end{subfigure}%
	\begin{subfigure}{.249\textwidth}
		\centering
		\includegraphics[width=0.99\linewidth]{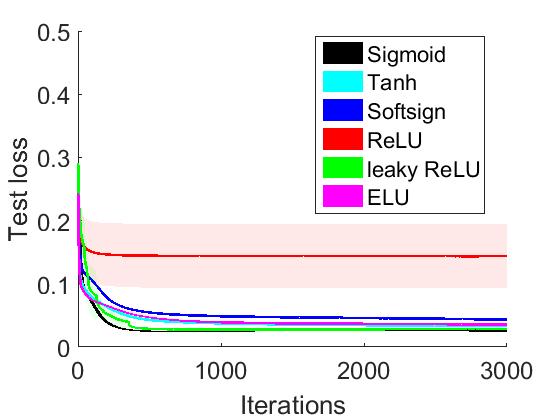}
	\end{subfigure}%
	\begin{subfigure}{.249\textwidth}
		\centering
		\includegraphics[width=0.99\linewidth]{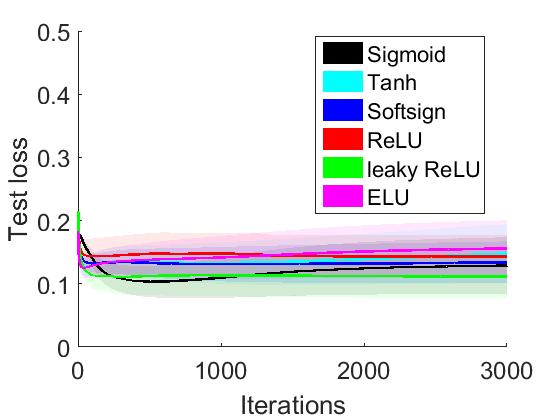}
	\end{subfigure}%
	
	\begin{subfigure}{.249\textwidth}
		\centering
		\includegraphics[width=0.99\linewidth]{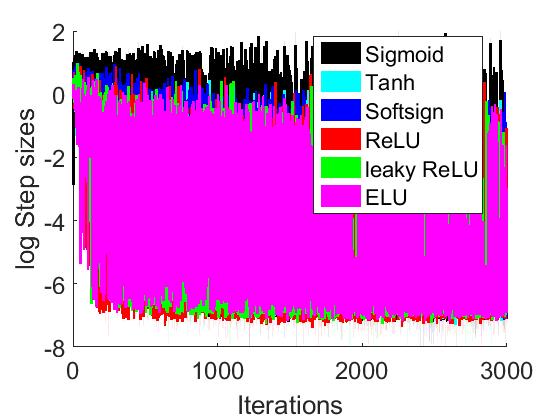}
		\caption{DS 1, $|\mathcal{B}_{n,i}|=64$}
	\end{subfigure}%
	\begin{subfigure}{.249\textwidth}
		\centering
		\includegraphics[width=0.99\linewidth]{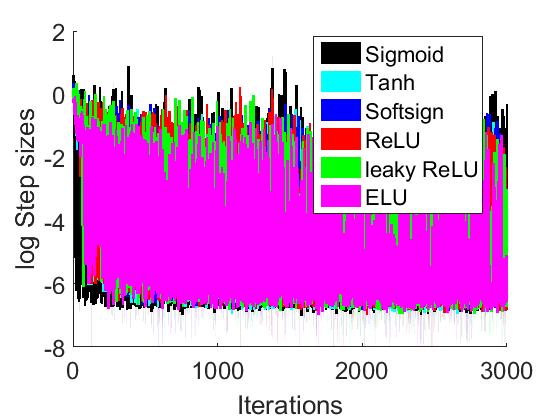}
		\caption{DS 2-4, $|\mathcal{B}_{n,i}|=64$}
	\end{subfigure}%
	\begin{subfigure}{.249\textwidth}
		\centering
		\includegraphics[width=0.99\linewidth]{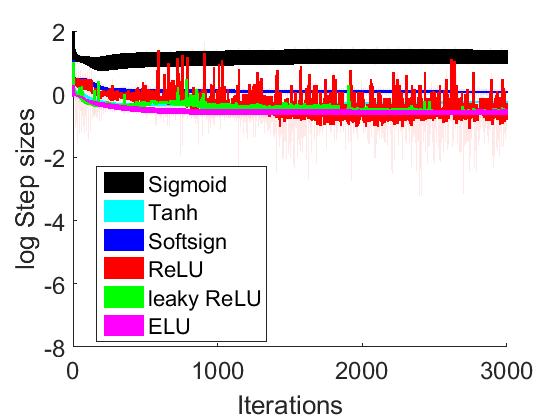}
		\caption{DS 1, $M$}
	\end{subfigure}%
	\begin{subfigure}{.249\textwidth}
		\centering
		\includegraphics[width=0.99\linewidth]{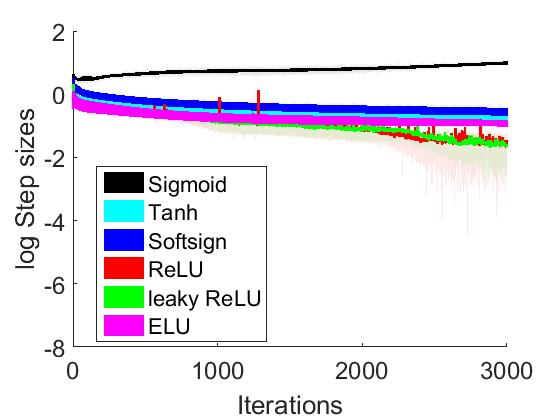}
		\caption{DS 2-4, $M$}
	\end{subfigure}%
	
	\caption{Training loss, test loss and the log of step sizes for a subset of datasets (DS) from the foundational problems dataset pool, trained using GOLS-B SGD (a,b) with mini-batch size $|\mathcal{B}_{n,i}|=64$ for investigation 5 and (c,d) using full-batches, $M$ in investigation 6. }
	\label{fig_foundational_fb}
\end{figure}

These considerations, as well as the results observed in Figure~\ref{fig_foundationals_mb}, suggest that the step sizes determined by GOLS-I are too large to result in stable training with ReLU. A contributing factor is that GOLS-I's initial accept condition \citep{Kafka2019jogo} allows univariate SNN-GPPs to be overshot in a controlled manner. Although overshoot has been shown to aid the training performance of LS-SGD \citep{Kafka2019jogo,Kafka2019}, it may be too aggressive to be implemented with ReLU AFs. Therefore, investigations 5 and 6 focus on whether GOLS-B \citep{Kafka2019jogo} with a conservative SNN-GPP bracketing strategy is capable of determining step sizes for LS-SGD in feedforward networks with ReLU AFs. The conservative bracketing strategy grows the minimum step size by a factor of 2, until a positive directional derivative is observed. This increases the probability of encountering SNN-GPPs in $B_\epsilon$ that are closest to $\boldsymbol{x}_n$ along the descent direction.

{
	In investigations 5 and 6, we more closely consider the loss functions of the first 4 foundational problems with ReLU AFs. The focus is primarily on the change in characteristics between dynamic MBSS and full-batch sampled losses. Therefore, we find the largest mini-batch size of the power 2 that allows MBSS for the first 4 problems. Due to separating problem data into training, validation and test datasets, the training dataset sizes for the first 4 problems are $M \in \{76, 108, 182, 263\}$ respectively. Therefore, the largest common mini-batch size of power 2 is $|\mathcal{B}|= 64$, which is implemented for investigation 5. Subsequently, investigation 6 uses full-batch sampling for the same datasets. The results for both investigations shown in Figure~\ref{fig_foundational_fb}, where all step sizes are determined using GOLS-B. Satisfied, that the validation and test datasets have been chosen representatively in Figure~\ref{fig_foundationals_mb}, we omit the validation dataset losses in Figure~\ref{fig_foundational_fb}. We also separate the results of dataset 1 in Figures~\ref{fig_foundational_fb}(a) and (c) from those of datasets 2-4 in Figures~\ref{fig_foundational_fb}(b) and (d), as these have distinct training characteristics. 
}


The results for training on dataset 1, which is the smallest considered dataset among the foundational problems, with $|\mathcal{B}_{n,i}|=64$ on the single hidden layer architecture with $2D$ nodes are shown in Figure~\ref{fig_foundational_fb}(a). Here, training the ReLU architecture is still unstable, while training the networks of the other AFs is effective. Note, that the step sizes of all AFs are significantly smaller and noisier with GOLS-B SGD compared to those of GOLS-I SGD. This is due to a combination of the conservative bracketing strategy and the lack of overshoot, compared to GOLS-I. However, this conservative approach aids in successfully training feedforward networks with the ReLU AF for datasets 2-4 in Figure~\ref{fig_foundational_fb}(b). This is confirmation, that the larger step sizes determined by GOLS-I led to the divergent training behaviour observed in Figure~\ref{fig_foundationals_mb}. However, this improved stability comes at the expense of training performance, as GOLS-B is an exact line search, which uses an order of magnitude more function evaluations per iteration compared to GOLS-I \citep{Kafka2019jogo}.


The full-batch sampled loss or true loss function results for investigation 6 are plotted in Figures~\ref{fig_foundational_fb}(c) and (d). The training performance of dataset 1 with ReLU in Figure~\ref{fig_foundational_fb}(c) shows little significant improvement in comparison to Figure~\ref{fig_foundational_fb}(a). The average training loss is only marginally better, with a mean drop in loss of $0.022$ for full-batch training. This indicates, that the deterministic loss function pertaining to dataset 1 with ReLU has descent directions leading into flat planes, which trap the training algorithm.

Generally, the determined step sizes for all AFs are significantly more stable for the full-batch case, than with $|\mathcal{B}_{n,i}|$. There are no incidences of minimum step sizes, as all search directions are deterministic descent directions. The variance that remains is due to the qualities of the deterministic descent direction itself, where the step size to the SNN-GPP along each descent direction is different. However, the step sizes of ReLU networks are particularly noisy, as GOLS-B SGD contends with the boundary between obtaining gradients that are dense, or sparse. Compared to dataset 1, this variance in step size is significantly reduced for datasets 2-4, which indicates that the boundary between obtaining dense and sparse gradient vectors along a descent direction is less abrupt, prompting less aggressive changes in step sizes. This is echoed by the training performance for ReLU with datasets 2-4, which is competitive with that of the remaining AFs. This suggests, that some ReLU feedforward network architectures construct small positive directional derivatives that "push" the line search back from zero-valued domains in the loss function, an observation confirmed by \cite{Kafka2019b}. 

An additional interesting observation between Figures~\ref{fig_foundational_fb}(b) and \ref{fig_foundational_fb}(d) is that the minimum test losses of most AFs are lower in the dynamic MBSS case, than when using full-batch training. It is clear, that training stagnates in Figure~\ref{fig_foundational_fb}(b) compared to Figure~\ref{fig_foundational_fb}(d), where the training loss decreases at a more rapid rate. However, as training slows in Figure~\ref{fig_foundational_fb}(b), the test losses for all AFs accept for Sigmoid are lower than their full-batch equivalent in Figure~\ref{fig_foundational_fb}(d). This indicates, that dynamic MBSS together with a conservative gradient-only line search can either slow training (as is the case for Sigmoid), or act as a regularizer during training (as is the case for all other AFs), which definitely warrants future study.

This investigation has demonstrated, that successfully determining step sizes using GOLS for feedforward neural networks with ReLU AFs is sensitive to both the line search strategy used, and the architecture of the given problem. In the examples shown, GOLS-B effectively resolved step sizes for LS-SGD in larger networks with ReLU AFs, while GOLS-I SGD was unable to conduct reliable training on the same problems. Step sizes that are too large, and variance produced by dynamic MBSS, lead to detailed features such as small positive directional derivatives being ignored. This impedes the ability of GOLS-I to determine effective step sizes for the ReLU AF. GOLS-B resolves more conservative step sizes, but bears a high computational cost. Instead, relaxing the absolute sparsity of ReLU, by implementing the leaky ReLU or ELU AF, significantly improves GOLS-I's ability to determine effective step sizes in dynamic MBSS losses for the feedforward architectures considered in this study. The non-zero derivatives of ELU and leaky ReLU in their negative input domains ensure that $\tilde{\boldsymbol{g}}(\boldsymbol{x})$ remains dense. This guarantees that all weights participate in update steps, and allows the training algorithm to recover from previous step sizes that were too large.


\subsection{MNIST with NetII}

\label{sec_NetII}

\begin{figure}[h!]
	\centering
	\begin{subfigure}{.33\textwidth}
		\centering 
		\includegraphics[width=0.99\linewidth]{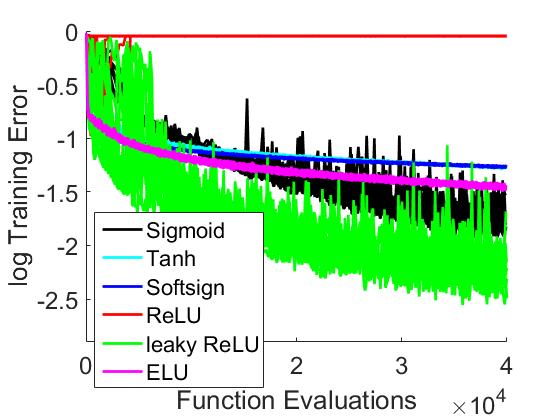}
		\caption{GOLS-I}
	\end{subfigure}%
	\begin{subfigure}{.33\textwidth}
		\centering
		\includegraphics[width=0.99\linewidth]{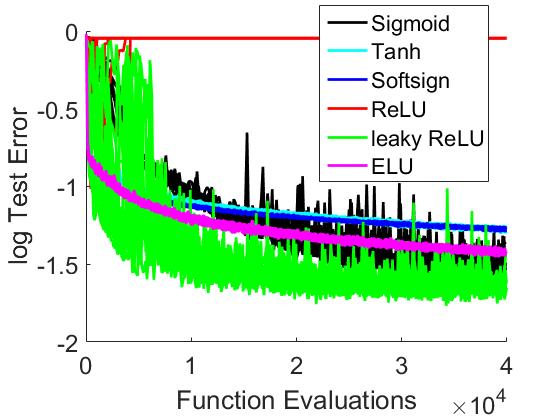}
		\caption{GOLS-I}
	\end{subfigure}%
	\begin{subfigure}{.33\textwidth}
		\centering
		\includegraphics[width=0.99\linewidth]{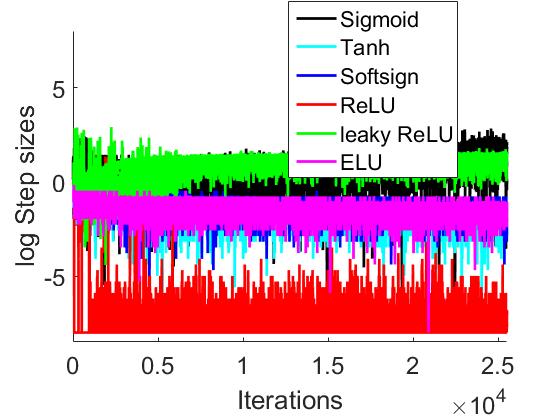}
		\caption{GOLS-I}
	\end{subfigure}%
	
	\begin{subfigure}{.33\textwidth}
		\centering 
		\includegraphics[width=0.99\linewidth]{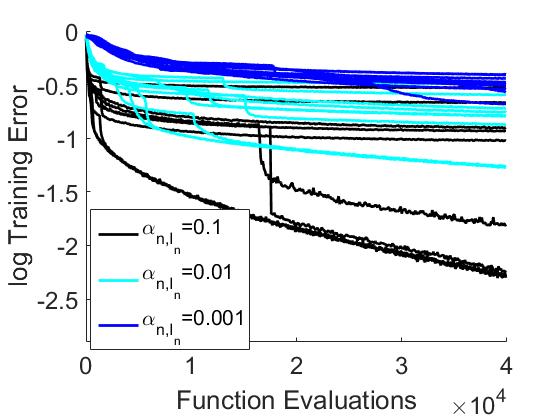}
		\caption{Fixed step sizes}
	\end{subfigure}%
	\begin{subfigure}{.33\textwidth}
		\centering
		\includegraphics[width=0.99\linewidth]{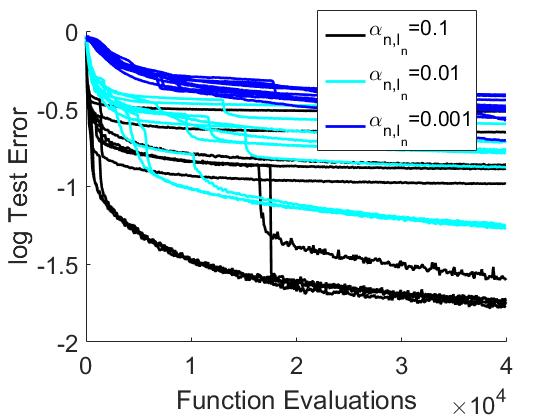}
		\caption{Fixed step sizes}
	\end{subfigure}%
	\begin{subfigure}{.33\textwidth}
		\centering
		\includegraphics[width=0.99\linewidth]{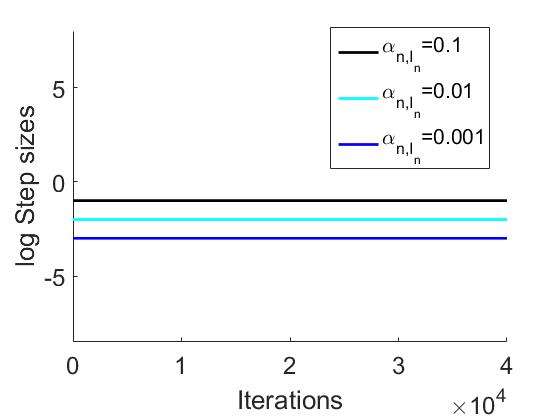}
		\caption{Fixed step sizes}
	\end{subfigure}%
	
	\begin{subfigure}{.33\textwidth}
		\centering 
		\includegraphics[width=0.99\linewidth]{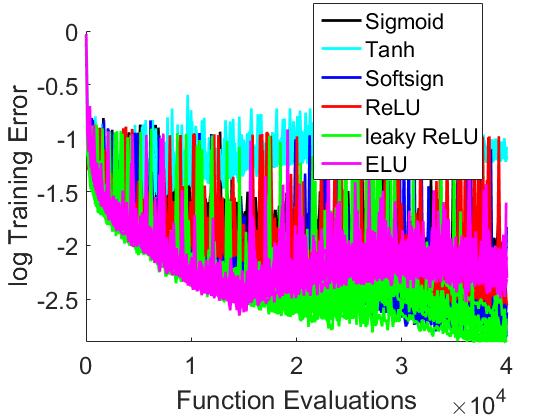}
		\caption{GOLS-I, batch-norm.}
	\end{subfigure}%
	\begin{subfigure}{.33\textwidth}
		\centering
		\includegraphics[width=0.99\linewidth]{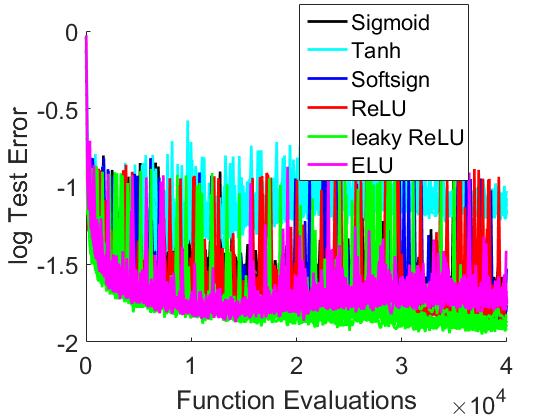}
		\caption{GOLS-I, batch-norm.}
	\end{subfigure}%
	\begin{subfigure}{.33\textwidth}
		\centering
		\includegraphics[width=0.99\linewidth]{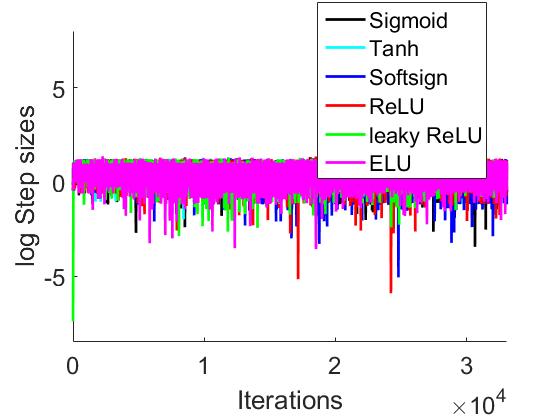}
		\caption{GOLS-I, batch-norm.}
	\end{subfigure}%
	
	\caption{Training and test classification errors with corresponding log of step sizes as obtained for the MNIST Dataset with the NetII architecture for different activation functions. Step sizes for LS-SGD are determined by (a)-(c) GOLS-I for the standard NetII architecture with all considered AFs, (d)-(f) a range of fixed step sizes for the standard NetII architecture with only ReLU, and (g)-(i) determined by GOLS-I for NetII with batch normalization and all considered AFs.}
	\label{fig_NIIm}
\end{figure}

Next, we present a larger training problem, where the choice of AF significantly influences training performance using GOLS-I SGD. The NetII architecture in combination with the well known MNIST dataset has been used to demonstrate the ability of probabilistic line searches to determine learning rates in dynamic MBSS loss functions \citep{Mahsereci2017a}. \citet{Mahsereci2017a} only implement the Tanh AF for this problem, which we extend by analysing all the AFs considered in Section~\ref{sec_connections} for investigation 7. In addition, we quantify the effect of batch normalization in investigation 9. The training and test classification errors, as well as accompanying step sizes for the different AFs are given in Figure~\ref{fig_NIIm}. We remind the reader, that the results presented for investigations 7-8 are given in function evaluations to be consistent with \citet{Mahsereci2017a}.

{
	In investigation 7, the training performance between different AFs varies significantly for the standard NetII architecture with GOLS-I SGD in Figures~\ref{fig_NIIm}(a)-(c). It is immediately evident, that ReLU is also unstable in the standard NetII architecture using GOLS-I SGD. The overall best performance is obtained using leaky ReLU, followed by Sigmoid. The next best performer is ELU, with Softsign marginally outperforming Tanh. Leaky ReLU trains particularly noisily in the first 5,000 function evaluations, before establishing a clear lead over the remaining AFs. Sigmoid trains slower than Tanh, Softsign and ELU during the first 10,000 function evaluations, but outperforms these thereafter. The relative performances of the AFs generalize, as they are also reflected in the test classification errors. There are a few indications that the resulting loss function of leaky ReLU and the Sigmoid AFs have different characteristics to those of Tanh, Softsign and ELU, namely: 1) The significantly lower training and test errors after 40,000 function evaluations, 2) the higher variance in error during training, and 3) step sizes that are up to two orders of magnitude larger than those of the rest. We speculate that this is due to specific interactions between activation function properties and the neural network architecture. 
	
	
	Both Sigmoid and leaky ReLU are AFs that "activate" in the positive input domain and tend towards zero in the negative input domain. Therefore, we postulate that the ability of the AFs to approximate zero function value outputs, while having non-zero derivatives, constructs loss function landscapes that are easier to traverse with GOLS-I SGD. This supports a study by \citet{Xu2016}, which found that a "penalized Tanh", that reduces the output magnitudes of function values and derivatives of Tanh in the negative input domain, performed competitively with sparsity class activation functions in deep convolutional neural network training. We postulate that the ability to significantly reduce the absolute function value of a node, decreases the information passed forward into a network. Subsequently, this reduces the sensitivity of nodes further downstream to the nodes that have low function value output. This contributes towards uncoupling parameters in the optimization space, $\boldsymbol{x}$, thus changing the nature of the loss function and resulting the unique training characteristics observed for Sigmoid and leaky ReLU.
}

Since ReLU demonstrates the same training difficulties with GOLS-I SGD as investigated in Section~\ref{sec_foundationalProbs}, we consider training ReLU using LS-SGD with fixed step sizes of $\alpha_{n,I_n} \in \{0.1,0.01,0.001\}$ for investigation 8 in Figures~\ref{fig_NIIm}(d)-(f). Again, training performance improves for ReLU using LS-SGD with fixed step sizes over GOLS-I SGD. However, the variance between each of the 10 training runs performed increases proportionately to the fixed step size. Compared to the relatively consistent performance of the other AFs with GOLS-I SGD, this result is unsatisfactory. Although individual training runs with $\alpha_{n,I_n} = 0.1$ outperform GOLS-I with Sigmoid or leaky ReLU, the fixed step size first needs to be determined, and subsequently multiple runs performed to obtain an appropriate $\alpha_{n,I_n}$. This highlights, that training using fixed step sizes is not a practical alternative to using GOLS-I with LS-SGD in feedforward neural networks and ReLU activation functions. As argued in Section~\ref{sec_foundationalProbs} implementing GOLS-B instead is computationally too demanding. It is therefore preferable to explore alternative means, by which the benefits of GOLS-I can be extended to ReLU architectures.

\begin{figure}[h!]
	\centering
	\begin{subfigure}{.33\textwidth}
		\centering
		\includegraphics[width=0.99\linewidth]{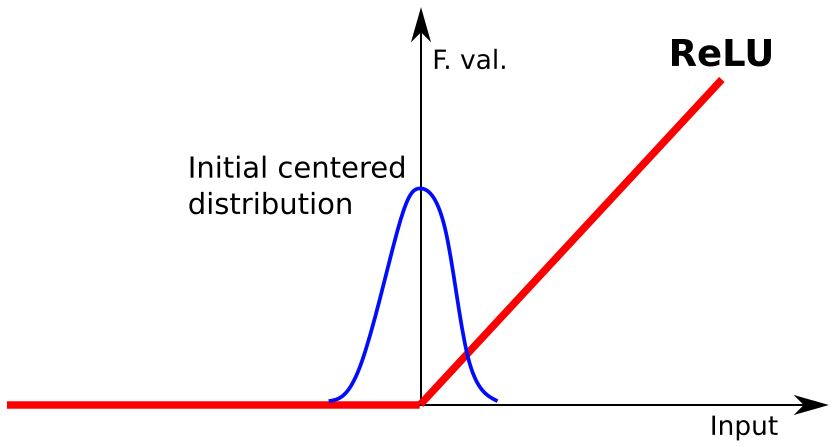}
		\caption{At initialization}
	\end{subfigure}%
	\begin{subfigure}{.33\textwidth}
		\centering
		\includegraphics[width=0.99\linewidth]{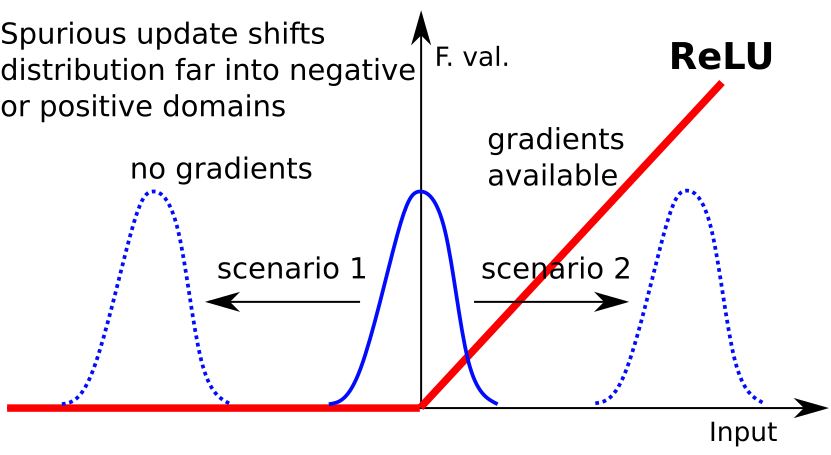}
		\caption{Spurious update w/o BN}
	\end{subfigure}%
	\begin{subfigure}{.33\textwidth}
		\centering
		\includegraphics[width=0.99\linewidth]{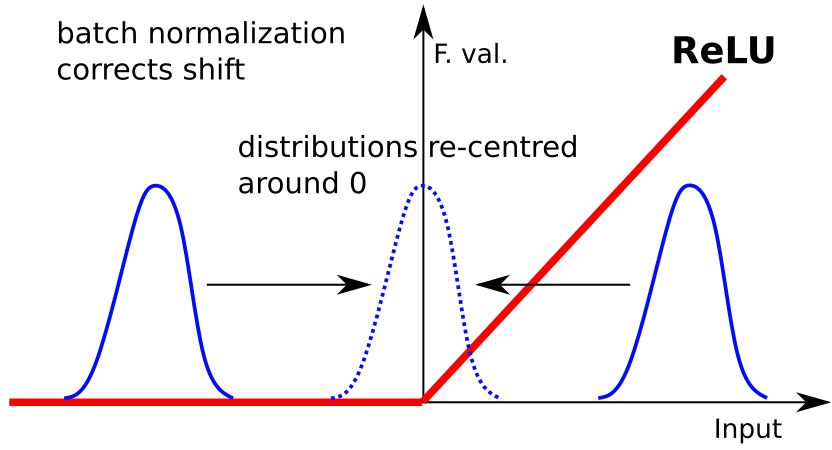}
		\caption{With batch normalization}
	\end{subfigure}%
	
	\caption{Schematic of the distribution of incoming information to a ReLU activation function (a) at initialization, (b) without batch normalization (BN) after spurious updates due to step sizes that are too large, and (c) after implementing batch normalization. Batch normalization centres the incoming information distribution around 0 and scales it, ensuring that the ReLU activation function remains active with a reasonable probability. Subsequently,  $\tilde{\boldsymbol{g}}(\boldsymbol{x})$ is more likely to remain dense than with the standard feedforward architecture.}
	\label{fig_relu_bn}
\end{figure}

The problem of training ReLU feedforward architectures using GOLS-I, as considered in Section~\ref{sec_foundationalProbs}, is summarized in Figure~\ref{fig_relu_bn}. At initialization, the distribution of information entering ReLU's input domain is centred around 0 \citep{He2016}, see Figure~\ref{fig_relu_bn}(a). This allows the "switching" mechanism proposed for ReLU to occur, whereby some data samples cause the node to "fire" (inputs in the positive domain) and others keep the node dormant (inputs in the negative domain). As discussed, large variance in gradient norms and the nature of a line search can cause step sizes that are spuriously too large, resulting in large changes in weight updates. Such updates can shift all the training dataset variance propagated through the network far into the negative or positive input domain of ReLU, see Figure~\ref{fig_relu_bn}(b). If all the data variance is in the positive input domain (scenario 2), gradients are available to allow subsequent update steps to correct for this shift. However, if all the dataset variance is in the negative input domain (scenario 1), the zero-derivative of ReLU prohibits information from travelling through the activation function to subsequent nodes. When this occurs to a significant portion of the network architecture, the flow of information through the network is significantly hampered and training stagnates.

Batch normalization \citep{Ioffe2015} is a method by which the inputs to a layer of nodes are continually centred by the mean of the previous layer's output and scaled by the corresponding variance. Applying batch normalization results in the distribution of information into a node remaining around the centre of the ReLU input domain, see Figure~\ref{fig_relu_bn}, increasing the likelihood of $\tilde{\boldsymbol{g}}(\boldsymbol{x})$ for the whole architecture remaining dense. This should increase the ability of GOLS-I to determine step sizes for ReLU, as partial derivatives are more likely to be non-zero, even after spurious updates. 

We implement batch normalization for NetII with the given AFs for investigation 9 and show the results in Figures~\ref{fig_NIIm}(g)-(i). It is clear, that the training performance of ReLU is drastically improved with GOLS-I. For the first time, it is possible to obtain competitive performance with ReLU using GOLS-I. Additionally, all activation functions show accelerated training with batch normalization over the standard NetII architecture. The best training performances are shared by Leaky ReLU and Softsign. However, the test errors are more comparable, for all AFs, ranging between $10^{-1.5}$ and $10^{-2}$ after 40,000 function evaluations. One exception is Tanh, which diverges after 5,000 function evaluations. However, the training and test errors achieved with batch normalization before 5,000 function evaluations are lower than those achieved for the standard architecture after 40,000 function evaluations. Therefore, although the reason for Tanh's divergence is worthy of further investigation, in this study we are satisfied with observing improved performance with Tanh due to batch normalization. 

It is noteworthy, that the step sizes for NetII with batch normalization have a consistent magnitude between different AFs and vary less in comparison to those determined for the standard NetII architecture. Since batch normalization alters the scaling of the search direction, the loss function seems more spherical to LS-SGD, which results in the step sizes being more consistent between AFs. However, this does not result in equal training performance between AFs (as seen for Tanh), which indicates that the different AFs still contribute unique characteristics to the loss function. Interestingly, the error variance characteristics also change, for many AFs. Leaky ReLU and Sigmoid errors remain noisy, but AFs such as Tanh, Softsign and ELU exhibit more variance with batch normalization than without, as their respective derivatives are highest around 0.


In summary, this investigation has shown that:

\begin{enumerate}
	\item Larger feedforward networks (than investigated in Section \ref{sec_foundationalProbs}) with ReLU AFs can also be unstable when training with GOLS-I.
	\item The interaction between neural network architecture and AF can lead to significant differences in training performance with GOLS-I (even when the systemically poor performance of ReLU is ignored).
	\item Implementing constant learning rates is not a viable alternative to determining step sizes with GOLS-I for ReLU AFs.
	\item Instead, batch normalization significantly improves training of ReLU architectures using GOLS-I.
	\item And lastly, batch normalization accelerated training for all AFs with GOLS-I in our investigation.
\end{enumerate}


\subsection{CIFAR10 with ResNet18}

Consider the interaction between architectures that use skip connections \citep{He2016} and different AFs using GOLS-I SGD. To this end, we implement the ResNet18 architecture, as applied to the CIFAR10 dataset. Skip connections ensure that the information flow to subsequent nodes remains unimpeded, regardless of whether an activation function such as ReLU prohibits information from travelling through a given node. The role of AFs in this case is to manipulate information that is additional to that transferred by the skip connections i.e. the "residuals". The interaction between the "skip-transferred" information and the residuals constructs the mapping behaviour between input and output domains of the neural network. The standard ResNet18 architecture includes batch normalization.

\begin{figure}[h!]
	\centering
	\begin{subfigure}{.33\textwidth}
		\centering
		\includegraphics[width=0.99\linewidth]{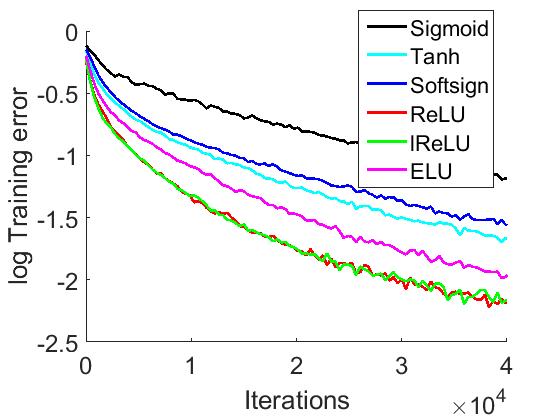}
		\caption{Standard}
	\end{subfigure}%
	\begin{subfigure}{.33\textwidth}
		\centering
		\includegraphics[width=0.99\linewidth]{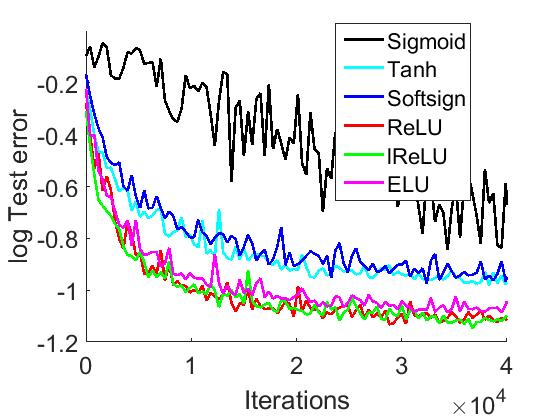}
		\caption{Standard}
	\end{subfigure}%
	\begin{subfigure}{.33\textwidth}
		\centering
		\includegraphics[width=0.99\linewidth]{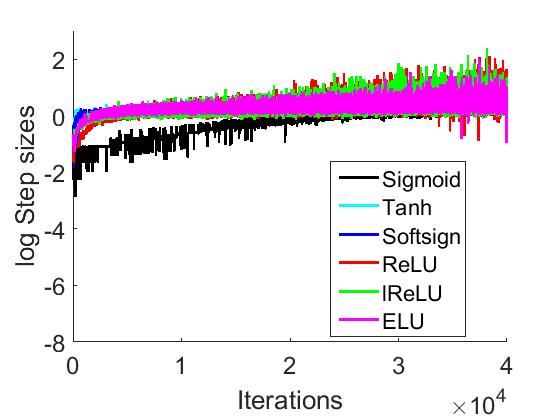}
		\caption{Standard}
	\end{subfigure}%
	
	\begin{subfigure}{.33\textwidth}
		\centering
		\includegraphics[width=0.99\linewidth]{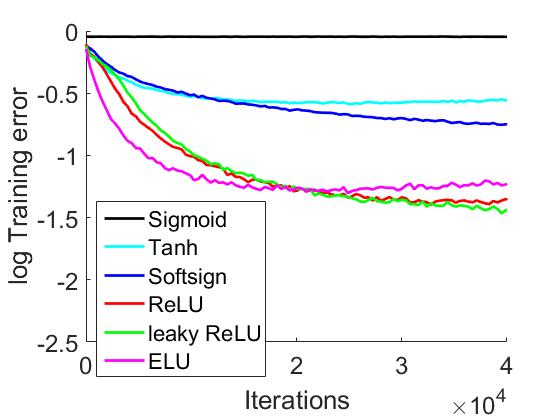}
		\caption{No Batch-Norm}
	\end{subfigure}%
	\begin{subfigure}{.33\textwidth}
		\centering
		\includegraphics[width=0.99\linewidth]{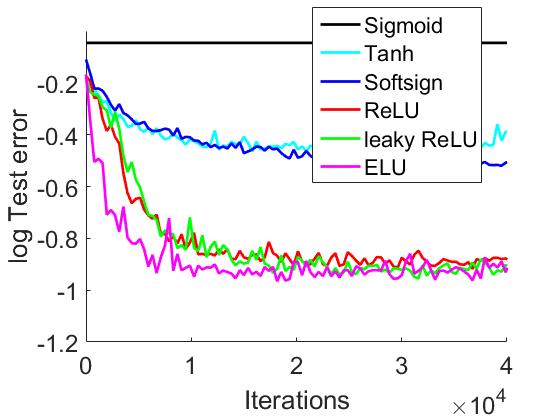}
		\caption{No Batch-Norm}
	\end{subfigure}%
	\begin{subfigure}{.33\textwidth}
		\centering
		\includegraphics[width=0.99\linewidth]{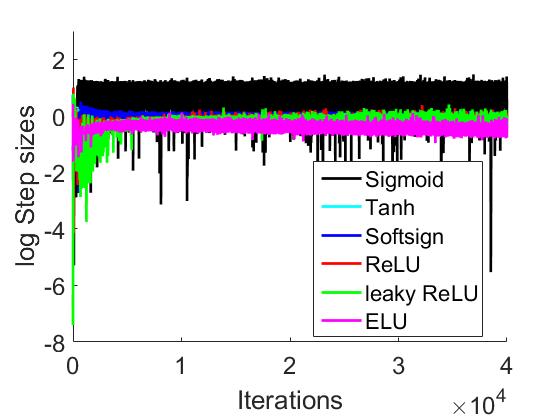}
		\caption{No Batch-Norm}
	\end{subfigure}%
	
	\caption{(a) Log training error, (b) log test error and (c) the log of step sizes for the CIFAR10 Dataset with the ResNet18 architecture trained using GOLS-I SGD. The standard ResNet architecture includes batch-norm layers, which are omitted in (d)-(f) in order to highlight the effect of skip connections on training with sparsity enforcing ReLU activation functions.}
	\label{fig_resnet}
\end{figure}

For investigation 10, we compare training and test classification error, as well as corresponding step sizes for ResNet18 with the considered range of AFs in Figures~\ref{fig_resnet}(a)-(c). In this case, there is a distinct difference in performance between the sparsity class and the saturation class of AFs. ReLU and leaky ReLU perform best in terms of training, with a slight advantage over ELU. However, this difference is less prominent in the test classification errors, where ELU is competitive with the rest of its class. A similar clustering occurs between Tanh and Softsign for the saturation class. Both perform very similarly in both training and test errors, with only a slight advantage belonging to Tanh. The Sigmoid AF is convincingly the worst performer of the considered AFs. Though training is slow and stable, the test losses are considerably noisier than those of the remaining AFs.

{
	Using Xavier initialization \citep{Glorot2010}, the initial weights for the Sigmoid AF are distributed around 0 in the input domain. However, an input distribution around zero corresponds to a function value output centred around 0.5 for Sigmoid. This is in contrast to the remaining saturation class AFs, which have outputs centred around 0. As shown by \cite{Glorot2010}, the cumulative effect of 0.5 outputs over a number of hidden layers can push Sigmoid AFs in later layers into saturation, where the derivative is significantly decreased. Batch normalization counters this problem by re-centring the outputs of network layers around 0, where the derivative is at a maximum. Additionally, the maximum derivative of Sigmoid is 0.25, which over many layers diminishes the gradient during backpropagation due to the chain-rule. Batch normalization also compensates for this property by scaling the variance to be 1 for every layer, ensuring that the gradient magnitudes remain adequately scaled. It is clear, that batch normalization has to do a significant amount of "correcting" for the Sigmoid AF in ResNet18. We suspect that the combination of these factors leads GOLS-I to estimate small initial step sizes for Sigmoid, with slow subsequent step size growth. However, the step sizes gradually increase to the point where they are comparable to those of the other AFs after 20,000 iterations. 
}

Apart from the step sizes of the Sigmoid AF, the step size magnitudes between the remaining AFs are similar and approximately constant. This matches the trend observed in Section~\ref{sec_NetII} for NetII with batch normalization. However, unlike the training performance in the NetII analysis, the Tanh AF does not diverge with ResNet18 and batch normalization. Interestingly, the magnitude and variance of determined step sizes for all AFs (except Sigmoid) increases significantly after 20,000 iterations. This correlates to an increase in gradient variance between data points associated with large losses (resulting in larger gradient norms), and those with lower losses (smaller gradient norms), as the architecture increasingly fits the data. thus, depending on the data-points in mini-batch, $\mathcal{B}_{n,i}$, the magnitude and direction of $\tilde{\boldsymbol{g}}(\boldsymbol{x})$ may vary significantly. Subsequently, this variance is transferred to the directional derivatives used by GOLS-I, leading to a higher variance in step sizes. A plausible corrective measure to manage this variance, is to gradually increase $\mathcal{B}_{n,i}$ as training progresses \citep{Friedlander2011,Smith2017}.

\begin{figure}[h!]
	\centering
	
	\begin{subfigure}{.49\textwidth}
		\centering
		\includegraphics[width=0.99\linewidth]{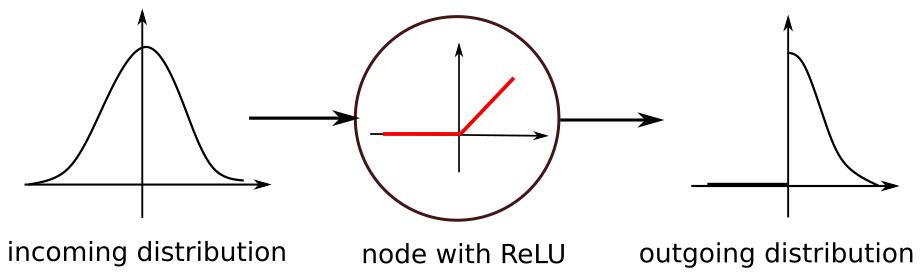}
		\caption{Standard Node}
	\end{subfigure}%
	
	\begin{subfigure}{.49\textwidth}
		\centering
		\includegraphics[width=0.99\linewidth]{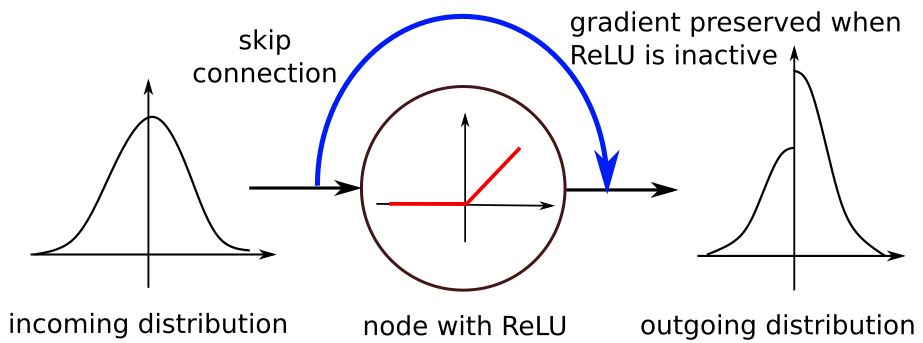}
		\caption{Node with skip-connection}
	\end{subfigure}%
	
	\caption{}
	\label{fig_AFmods}
\end{figure}

{
	Since ResNet is constructed with both skip-connections and batch normalization, it is unclear, which of the two architectural features contribute more significantly towards improving training performance with ReLU. We have considered the contribution of batch normalization to improving training in Section~\ref{sec_NetII}. Here, we consider skip connections more closely. The difference between standard feedforward nodes and skip connected nodes is illustrated in Figure~\ref{fig_AFmods} with the ReLU AF. A standard node, shown in Figure~\ref{fig_AFmods}(a) passes the incoming distribution through the activation function at the node, which augments the distribution according to its characteristics. As demonstrated in Figure~\ref{fig_relu_bn}, this can be problematic with the ReLU AF if poor updates occur, as the propagation of gradients can become obstructed. For a skip connected node, shown in Figure~\ref{fig_AFmods}(b), the incoming distribution is added to the output of the standard node. This ensures, that even in the worst case, where no information passes through the AF, the incoming distribution always propagates forward. Subsequently, evaluated gradients will always be dense.
	
	In investigation 11 we observe the influence of skip connections in ResNet18. Therefore, investigation 11 repeats the analysis of investigation 10, albeit without the use of batch normalization. The results are shown in Figures~\ref{fig_resnet}(d)-(f). As is consistent with investigations 7 and 9 performed in Section~\ref{sec_NetII}, training performance slows without the use of batch normalization for all AFs. The results of investigation 11 show that training progresses for all AFs, with the exception of the Sigmoid AF. We postulate that this drop in performance is due to scaling difficulties in deep neural networks driven by Sigmoid's positive offset, small maximum derivative and saturating nature \citep{Glorot2010}, which are subsequently not corrected by batch normalization. 
	
	We confirmed this phenomenon by conducting numerous additional runs with modified versions of AFs considered in this study. A modified version of the Tanh AF, namely $0.5\cdot$Tanh$+0.5$ which is centred around 0.5 and has a maximum derivative of 0.5, performed the same as Sigmoid. Conversely, successful training was observed using $2\cdot$Sigmoid$-1$, which also has a maximum derivative of 0.5 while being centred around 0. This suggests that passing through the origin is an important AF property for promoting effective training with deep networks, an assumption held for both Xavier \citep{Glorot2010} and He \citep{He2015} initialization strategies. Additionally, a modified leaky ReLU with maximum output derivative $< 0.5$ also failed to train. Indeed, Xavier initialization assumes that AF derivatives are 1 around the zero input domain \citep{Glorot2010}. As Sigmoid satisfies neither of these properties, it is not surprising that its implementation in ResNet18 without batch normalization failed to train successfully.
}

Importantly, training ResNet18 without bath normalization with the ReLU AF using GOLS-I SGD is not only stable, it also shares the best training performance with leaky ReLU. This is significant, as it demonstrates the effectiveness of skip connections in ensuring that gradients remain dense with ReLU AFs. This allows ReLU's performance to be directly compared to that of the remaining AFs without batch-normalization. Both ReLU and leaky ReLU are outperformed by ELU during the first half of training, but overtake it during the latter half. Figures~\ref{fig_resnet}(d) and (e) suggest that within the context of skip connections, the difference in performance between ReLU and Leaky ReLU is insignificant.

It is an emerging trend, that the sparsity class of AFs gradually begins to outperform the saturation class as the size of considered neural networks increases. For the foundational problems of Section~\ref{sec_foundationalProbs}, this difference is marginal, as the number of nodes in the architecture increases. For the NetII problem in Section~\ref{sec_NetII}, two of the top three performers are from the sparsity class for architectures with and without batch normalization. In the ResNet18 architecture, all of the sparsity class AFs outperform all of the saturation class AFs both with and without batch normalization. This suggests, that the sparsity property becomes increasingly useful, as the size of the network increases. This is consistent with how sparsity operates, as the number of "channels" available to construct the mapping between input and output spaces increases with growing architecture size.

This investigation demonstrates that skip-connections are effective in ensuring that $\tilde{\boldsymbol{g}}(\boldsymbol{x})$ remains dense for ReLU AFs, where sparsity is enforced. In cases where the outputs of ReLU nodes are zero, it is only the residual that is zero, while the information of previous nodes is still passed to subsequent layers through skip connections. This drastically improves GOLS-I SGD's ability to train ReLU neural network architectures. Subsequently, batch normalization layers contribute additional benefit for all AFs considered in our investigations, by increasing robustness and accelerating training. Additionally, this investigation supports the trends observed in Sections~\ref{sec_foundationalProbs} and \ref{sec_NetII}, where the training performance of larger architectures using GOLS-I SGD is improved by implementing sparsity class AFs.

\section{Conclusion}


In this study, we consider the interaction between gradient-only line searches and a variety of neural networks constructed with six different activation functions. The activation functions considered are split into two classes, namely the saturation class (Sigmoid, Tanh and Softsign), and the sparsity class (ReLU, leaky ReLU and ELU). Gradient-only line searches are used to determine the step sizes for gradient based optimizers in full-batch and dynamic mini-batch sub-sampled (MBSS) loss functions. In our study we implement the gradient-only line search that is inexact (GOLS-I) as well as the Gradient-Only Line Search with Bisection (GOLS-B) \citep{Kafka2019jogo}. Eleven investigations are conducted using 13 different datasets with a total of 37 network architectures, some using the cross entropy loss and others the mean squared error loss. Training problems include 11 foundational datasets with feedforward neural networks, the MNIST dataset with NetII \citep{Mahsereci2017a} and the CIFAR10 dataset with ResNet18 \citep{He2016}. These problems cover a range of architectural features, including batch normalization \citep{Ioffe2015} and skip connections. 

We find, that GOLS is effective in determining the step sizes in dynamic MBSS training for all but a few combinations of activation functions and network architectures. The small neural networks show a close grouping in training performance between the considered activation functions, with a slight advantage belonging to non-linear, saturation type activations. However, training performance of feedforward networks with the ReLU activation function, coupled with GOLS performed poorly. Analyses with NetII and ResNet18 without batch normalization show that a particular activation function can significantly improve the training performance for a given network architecture with GOLS-I. For NetII, the best performers were leaky ReLU and Sigmoid activation functions, while the troublesome performance of ReLU with GOLS-I seen in the foundational problems recurred for NetII.

Our investigations suggest that the predominant cause of GOLS-I's inability to train ReLU architectures are weight updates with step sizes that are too large. These updates shift the distribution of information entering a ReLU activation function fully into its inactive domain, thus producing zero-outputs and halting the flow of information through a node. This can lead to large portions of the network becoming and remaining inactive, which leads to the gradient vector being sparse in these cases. In addition, the implication that there exist domains in the loss function that have zero-gradients means that ReLU loss functions with feedforward architectures can break the assumptions of Lyapunov's global stability theorem, which govern the convergence properties of GOLS-I.

However, the training difficulties encountered with ReLU architectures using GOLS-I can be alleviated by implementation of batch normalization, and skip connections as used in residual networks. Batch normalization centres the distribution of information flowing between sequential network layers around zero, increasing the number of active nodes in the network. Alternatively, skip connections by design guarantee the propagation of input information throughout the network architecture. These technologies ensure that the gradient vector remains dense, allowing GOLS-I to recover form spurious updates, when they occur. Both skip connections and batch normalization render ReLU's competitive with the other activation functions. Batch normalization has the added benefit of accelerating training for all activation functions. Training ResNet18 using GOLS-I with and without batch normalization demonstrated, that additionally implementing batch normalization with skip connections results in a double benefit of stability for ReLU, as well as accelerated training for all activation functions.

Gradient-only line searches are effective at determining adaptive step sizes for gradient descent based training algorithms. Our studies have demonstrated, that the interaction between activation functions and neural network architectures matter. The ResNet18 training problem showed a clear distinction between saturation and sparsity classes of activation functions, with superior training performance belonging to the latter group. The properties of an activation function's derivative have a direct effect on the nature of the loss function presented to the optimization algorithm. Our studies suggest, that activation functions that promote sparsity are better suited to larger network architectures than classical saturation type activation functions. Additionally, significant difficulties can be encountered when training feedforward ReLU architectures with GOLS-I. Therefore, we suggest that practitioners consider technologies such as batch normalization and skip connections, when constructing neural network architectures to be trained with GOLS-I.

\section*{Acknowledgements}
This work was supported by the Centre for Asset and Integrity Management (C-AIM), Department of Mechanical and Aeronautical Engineering, University of Pretoria, Pretoria, South Africa. We would also like to thank NVIDIA for sponsoring the Titan X Pascal GPU used in this study.






\begin{thebibliography}{}
	
	\bibitem[Arora, 2011]{Arora2011}
	Arora, J. (2011).
	\newblock {\em {Introduction to Optimum Design, Third Edition}}.
	\newblock Academic Press Inc.
	
	\bibitem[Bergstra et~al., 2009]{Bergstra2009}
	Bergstra, J., Desjardins, G., Lamblin, P., and Bengio, Y. (2009).
	\newblock {Quadratic Polynomials Learn Better Image Features}.
	\newblock In {\em Technical Report 1337, IT Department and operations research,
		University of Montreal}, pages 1--11.
	
	\bibitem[Bollapragada et~al., 2017]{Bollapragada2017}
	Bollapragada, R., Byrd, R., and Nocedal, J. (2017).
	\newblock {Adaptive Sampling Strategies for Stochastic Optimization}.
	\newblock {\em arXiv:1710.11258}, pages 1--32.
	
	\bibitem[Bottou, 2010]{Bottou2010}
	Bottou, L. (2010).
	\newblock {Large-Scale Machine Learning with Stochastic Gradient Descent}.
	\newblock In {\em COMPSTAT 2010, Keynote, Invited and Contributed Papers},
	volume~19, pages 177--186.
	
	\bibitem[Boyd et~al., 2003]{Boyd2003}
	Boyd, S., Xiao, L., and Mutapcic, A. (2003).
	\newblock {Subgradient Methods}.
	\newblock In {\em lecture notes of EE392o, Stanford University}, volume~1,
	pages 1--21.
	
	\bibitem[{Chae} and {Wilke}, 2019]{Chae2019}
	{Chae}, Y. and {Wilke}, D.~N. (2019).
	\newblock Empirical study towards understanding line search approximations for
	training neural networks.
	\newblock {\em arXiv:1909.06893 [stat.ML]}, pages 1--30.
	
	\bibitem[Choromanska et~al., 2015]{Choromanska2015}
	Choromanska, A., Henaff, M., Mathieu, M., Arous, G.~B., and LeCun, Y. (2015).
	\newblock {The Loss Surfaces of Multilayer Networks}.
	\newblock In {\em AISTATS 2015}, volume~38, pages 192--204.
	
	\bibitem[Clevert et~al., 2016]{Clevert2016}
	Clevert, D.-A., Unterthiner, T., and Hochreiter, S. (2016).
	\newblock {Fast and Accurate Deep Network Learning by Exponential Linear Units
		(ELUS)}.
	\newblock In {\em ICLR 2016}, pages 1--14.
	
	\bibitem[Dauphin et~al., 2014]{Dauphin2014}
	Dauphin, Y., Pascanu, R., Gulcehre, C., Cho, K., Ganguli, S., and Bengio, Y.
	(2014).
	\newblock {Identifying and attacking the Saddle Point Problem in
		High-Dimensional Non-Convex Optimization}.
	\newblock In {\em ICLR 2014}, pages 1--9.
	
	\bibitem[Duchi et~al., 2011]{Duchi2011}
	Duchi, J., Hazan, E., and Singer, Y. (2011).
	\newblock {Adaptive Subgradient Methods for Online Learning and Stochastic
		Optimization}.
	\newblock {\em Journal of Machine Learning Research}, 12(July):2121--2159.
	
	\bibitem[Fisher, 1936]{Fisher1936}
	Fisher, R.~A. (1936).
	\newblock {The use of Multiple Measurements in Taxonomic Problems}.
	\newblock {\em Annals of Eugenics}, 7(2):179--188.
	
	\bibitem[Friedlander and Schmidt, 2011]{Friedlander2011}
	Friedlander, M.~P. and Schmidt, M. (2011).
	\newblock {Hybrid Deterministic-Stochastic Methods for Data Fitting}.
	\newblock {\em arXiv:1104.2373 [cs.LG]}, pages 1--26.
	
	\bibitem[Glorot and Bengio, 2010]{Glorot2010}
	Glorot, X. and Bengio, Y. (2010).
	\newblock {Understanding the Difficulty of Training Deep Feedforward Neural
		Networks}.
	\newblock {\em arXiv:1308.0850}, pages 1--8.
	
	\bibitem[Glorot and Bordes, 2011]{Glorot2011}
	Glorot, X. and Bordes, A. (2011).
	\newblock {Deep Sparse Rectifier Neural Networks}.
	\newblock In {\em Proceedings of Machine Learning Research}, volume~15, pages
	315--323.
	
	\bibitem[Goodfellow et~al., 2015]{Goodfellow2015}
	Goodfellow, I.~J., Vinyals, O., and Saxe, A.~M. (2015).
	\newblock {Qualitatively Characterizing Neural Network Optimization Problems}.
	\newblock In {\em ICLR 2015}, pages 1--11.
	
	\bibitem[Han and Morag, 1995]{Han1995}
	Han, J. and Morag, C. (1995).
	\newblock {The Influence of the Sigmoid Function Parameters on the Speed of
		Backpropagation Learning}.
	\newblock In Mira, J. and Sandoval, F., editors, {\em Natural to Artificial
		Neural Computation. Lecture Notes in Computer Science}, volume 930. Springer.
	
	\bibitem[{He} et~al., 2015]{He2015}
	{He}, K., {Zhang}, X., {Ren}, S., and {Sun}, J. (2015).
	\newblock Delving deep into rectifiers: Surpassing human-level performance on
	imagenet classification.
	\newblock {\em arXiv:1502.01852 [cs.CV]}, pages 1--11.
	
	\bibitem[He et~al., 2016]{He2016}
	He, K., Zhang, X., Ren, S., and Sun, J. (2016).
	\newblock {Deep Residual Learning for Image Recognition}.
	\newblock In {\em IEEE 2016 IEEE Conference on Computer Vision and Pattern
		Recognition (CVPR) - Las Vegas, NV, USA (2016.6.27-2016.6.30)}.
	
	\bibitem[{Ioffe} and {Szegedy}, 2015]{Ioffe2015}
	{Ioffe}, S. and {Szegedy}, C. (2015).
	\newblock Batch normalization: Accelerating deep network training by reducing
	internal covariate shift.
	\newblock {\em arXiv:1502.03167 [cs.LG]}, pages 1--9.
	
	\bibitem[Johnson et~al., 2012]{Johnson2012}
	Johnson, B., Tateishi, R., and Xie, Z. (2012).
	\newblock {Using Geographically Weighted Variables for Image Classification}.
	\newblock {\em Remote Sensing Letters}, 3(6):491--499.
	
	\bibitem[{Kafka} and {Wilke}, 2019]{Kafka2019}
	{Kafka}, D. and {Wilke}, D.~N. (2019).
	\newblock {Gradient-Only Line Searches: An Alternative to Probabilistic Line
		Searches}.
	\newblock {\em arXiv:1903.09383 [stat.ML]}, pages 1--25.
	
	\bibitem[Kafka and Wilke, 2019a]{Kafka2019jogo}
	Kafka, D. and Wilke, D.~N. (2019a).
	\newblock {Resolving Learning Rates Adaptively by locating Stochastic
		Non-Negative Associated Gradient Projection Points using Line Searches}.
	\newblock Unpublished: In review at the Journal of Global Optimization.
	
	\bibitem[Kafka and Wilke, 2019b]{Kafka2019b}
	Kafka, D. and Wilke, D.~N. (2019b).
	\newblock {Visual Interpretation of the Robustness of Non-Negative Associative
		Gradient Projection Points over Function Minimizers in Mini-Batch Sampled
		Loss Functions}.
	\newblock {\em arXiv:1903.08552 [stat.ML]}, pages 1--32.
	
	\bibitem[Karlik, 2015]{Karlik2015}
	Karlik, B. (2015).
	\newblock {Performance Analysis of Various Activation Functions in Generalized
		MLP Architectures of Neural Networks}.
	\newblock {\em International Journal of Artificial Intelligence and Expert
		Systems}, 1(4):111--122.
	
	\bibitem[Krizhevsky and Hinton, 2009]{Krizhevsky2009}
	Krizhevsky, A. and Hinton, G.~E. (2009).
	\newblock {Learning Multiple Layers of Features from Tiny Images}.
	\newblock {\em University of Toronto}.
	
	\bibitem[Krizhevsky et~al., 2012]{Krizhevsky2012}
	Krizhevsky, A., Sutskever, I., and Hinton, G.~E. (2012).
	\newblock {ImageNet Classification with Deep Convolutional Neural Networks}.
	\newblock In {\em NIPS 2012}, pages 1--9.
	
	\bibitem[{Kungurtsev} and {Pevny}, 2018]{Kungurtsev2018}
	{Kungurtsev}, V. and {Pevny}, T. (2018).
	\newblock {Algorithms for solving optimization problems arising from deep
		neural net models: smooth problems}.
	\newblock {\em arXiv:1807.00172 [math.OC]}, pages 1--5.
	
	\bibitem[Lecun et~al., 1998]{Lecun1998}
	Lecun, Y., Bottou, L., Bengio, Y., and Haffner, P. (1998).
	\newblock {Gradient-Based Learning Applied to Document Recognition}.
	\newblock {\em Proceedings of the IEEE}, 86(11):2278--2324.
	
	\bibitem[Li et~al., 2017]{Li2017}
	Li, H., Xu, Z., Taylor, G., Studer, C., and Goldstein, T. (2017).
	\newblock {Visualizing the Loss Landscape of Neural Nets}.
	\newblock {\em arXiv:1712.09913}, pages 1--21.
	
	\bibitem[Liu, 2018]{Kuangliu2018}
	Liu, K. (2018).
	\newblock {95.16\% on CIFAR10 with PyTorch}.
	\newblock \url{https://github.com/kuangliu/pytorch-cifar}.
	\newblock Accessed: 2018-09-30.

	\bibitem[Liu et~al., 2017]{Liu2017}
	Liu, P.,  Zeng, Z., Wang, J. (2017).
	\newblock {Multistability of Delayed Recurrent Neural Networks with Mexican Hat Activation Functions}.
	\newblock {\em Neural Computation}, 29(2):423--457.
	
	\bibitem[Lucas et~al., 2013]{Lucas2013}
	Lucas, D.~D., Klein, R., Tannahill, J., Ivanova, D., Brandon, S., Domyancic,
	D., and Zhang, Y. (2013).
	\newblock {Failure Analysis of Parameter-Induced Simulation Crashes in Climate
		Models}.
	\newblock {\em Geoscientific Model Development}, 6(4):1157--1171.
	
	\bibitem[Lyapunov, 1992]{Lyapunov1992}
	Lyapunov, A.~M. (1992).
	\newblock {The General Problem of the Stability of Motion}.
	\newblock {\em International Journal of Control}, 55(3):531--534.
	
	\bibitem[Maas et~al., 2013]{Maas2013}
	Maas, A.~L., Hannun, A.~Y., and Ng, A.~Y. (2013).
	\newblock {Rectifier Nonlinearities improve Neural Network Acoustic Models}.
	\newblock In {\em ICML 2013}, volume~28, page~6.
	
	\bibitem[Mahsereci and Hennig, 2017]{Mahsereci2017a}
	Mahsereci, M. and Hennig, P. (2017).
	\newblock {Probabilistic Line Searches for Stochastic Optimization}.
	\newblock {\em Journal of Machine Learning Research}, 18:1--59.
	
	\bibitem[Mansouri et~al., 2013]{Mansouri2013}
	Mansouri, K., Ringsted, T., Ballabio, D., Todeschini, R., and Consonni, V.
	(2013).
	\newblock {Quantitative Structure–Activity Relationship Models for Ready
		Biodegradability of Chemicals}.
	\newblock {\em Journal of Chemical Information and Modeling}, 53(4):867--878.
	
	\bibitem[Prechelt, 1994]{Prechelt1994}
	Prechelt, L. (1994).
	\newblock {PROBEN1 - a set of neural network benchmark problems and
		benchmarking rules (Technical Report 21-94)}.
	\newblock Technical report, Universit{\"{a}}t Karlsruhe.
	
	\bibitem[pytorch.org, 2019]{PyTorch1}
	pytorch.org (2019).
	\newblock {PyTorch}.
	\newblock \url{https://pytorch.org/}.
	\newblock Version: 1.0.
	
	\bibitem[Robbins and Monro, 1951]{Robbins1951}
	Robbins, H. and Monro, S. (1951).
	\newblock {A Stochastic Approximation Method}.
	\newblock {\em The Annals of Mathematical Statistics}, 22(3):400--407.
	
	\bibitem[Saxe et~al., 2013]{Saxe2013}
	Saxe, A.~M., McClelland, J.~L., and Ganguli, S. (2013).
	\newblock {Exact Solutions to the Nonlinear Dynamics of Learning in Deep Linear
		Neural Networks}.
	\newblock {\em CoRR}, abs/1312.6:1--22.
	
	\bibitem[Schraudolph, 1999]{Schraudolph1999}
	Schraudolph, N.~N. (1999).
	\newblock {Local Gain Adaptation in Stochastic Gradient Descent}.
	\newblock {\em 9th International Conference on Artificial Neural Networks:
		ICANN '99}, 1999:569--574.
	
	\bibitem[Schraudolph and Graepel, 2003]{Schraudolph2003}
	Schraudolph, N.~N. and Graepel, T. (2003).
	\newblock {Combining conjugate direction methods with stochastic approximation
		of gradients}.
	\newblock {\em Proceedings of the Ninth International Workshop on Artificial
		Intelligence and Statistics, AISTATS 2003}, pages 2--7.
	
	\bibitem[Schraudolph et~al., 2007]{Schraudolph2006}
	Schraudolph, N.~N., Yu, J., and G{\"{u}}nter, S. (2007).
	\newblock {A Stochastic Quasi-Newton Method for Online Convex Optimization}.
	\newblock {\em International Conference on Artificial Intelligence and
		Statistics}, pages 436----443.
	
	\bibitem[Smith, 2015]{Smith2015}
	Smith, L.~N. (2015).
	\newblock {Cyclical Learning Rates for Training Neural Networks}.
	\newblock {\em arXiv:1506.01186}.
	
	\bibitem[Smith et~al., 2017]{Smith2017}
	Smith, S.~L., Kindermans, P.-J., Ying, C., and Le, Q.~V. (2017).
	\newblock {Don't Decay the Learning Rate, Increase the Batch Size}.
	\newblock {\em arXiv:1711.00489}.
	
	\bibitem[Snyman and Wilke, 2018]{Snyman2018}
	Snyman, J.~A. and Wilke, D.~N. (2018).
	\newblock {\em {Practical Mathematical Optimization}}, volume 133 of {\em
		Springer Optimization and Its Applications}.
	\newblock Springer International Publishing, Cham.
	
	\bibitem[Tong and Liu, 2005]{Tong2005}
	Tong, F. and Liu, X. (2005).
	\newblock {Samples Selection for Artificial Neural Network Training in
		Preliminary Structural Design}.
	\newblock {\em Tsinghua Science {\&} Technology}, 10(2):233--239.
	
	\bibitem[Werbos, 1994]{Werbos1994}
	Werbos, P.~J. (1994).
	\newblock {\em {The Roots of Backpropagation: From Ordered Derivatives to
			Neural Networks and Political Forecasting}}.
	\newblock Wiley-Interscience, New York, NY, USA.
	
	\bibitem[Wilke et~al., 2013]{Wilke2013}
	Wilke, D.~N., Kok, S., Snyman, J.~A., and Groenwold, A.~A. (2013).
	\newblock {Gradient-Only Approaches to Avoid Spurious Local Minima in
		Unconstrained Optimization}.
	\newblock {\em Optimization and Engineering}, 14(2):275--304.
	
	\bibitem[Wilson and Martinez, 2003]{Wilson2003}
	Wilson, D.~R. and Martinez, T.~R. (2003).
	\newblock {The General Inefficiency of Batch Training for Gradient Descent
		Learning}.
	\newblock {\em Neural Networks}, 16(10):1429--1451.
	
	\bibitem[{Xu} et~al., 2016]{Xu2016}
	{Xu}, B., {Huang}, R., and {Li}, M. (2016).
	\newblock {Revise Saturated Activation Functions}.
	\newblock {\em arXiv:1602.05980[cs.LG]}, pages 1--7.
	
\end{thebibliography}


%
%
%

\end{document}